\pdfoutput=1
\documentclass[11pt]{article}

\usepackage[final]{acl}
\usepackage{times}
\usepackage{latexsym}

\usepackage[T1]{fontenc}

\usepackage[utf8]{inputenc}

\usepackage{microtype}

\usepackage{inconsolata}

\newcommand{\highlight}[1]{'#1'}
\usepackage{graphicx}      
\usepackage{float}         
\usepackage{caption}       
\usepackage{subcaption}    

\usepackage{booktabs}
\usepackage{tabularx}
\usepackage{makecell} 
\usepackage{multirow} 

\usepackage{listings}  
\usepackage{xcolor}    

\lstdefinelanguage{Prompts}{
    morekeywords={system, user},
    sensitive=false,
    morecomment=[l]{//},
    morestring=[b]"
}

\lstdefinestyle{promptstyle}{
    language=Prompts,
    basicstyle=\ttfamily\small,
    keywordstyle=\bfseries\color{black},
    stringstyle=\color{brown},
    commentstyle=\color{gray},
    breaklines=true,
    breakindent=0pt,
    breakautoindent=false,
    showstringspaces=false,
    frame=single,
    rulecolor=\color{black},  
    backgroundcolor=\color{lightgray!20},
    xleftmargin=1em,
    xrightmargin=1em,
    captionpos=b
}

\title{Analyzing the Role of Context in Forecasting with Large Language Models}

\author{Gerrit Mutschlechner \\
  University of Innsbruck \\
  Innsbruck, Austria \\
  \texttt{gerrit.mutschlechner@student.uibk.ac.at} \\\And
  Adam Jatowt \\
  University of Innsbruck \\
  Innsbruck, Austria \\
  \texttt{adam.jatowt@uibk.ac.at}}

\begin{document}
\maketitle
\begin{abstract}

This study evaluates the forecasting performance of recent language models (LLMs) on binary forecasting questions. We first introduce a novel dataset of over 600 binary forecasting questions, augmented with related news articles and their concise question-related summaries\footnote{The dataset can be obtained after contacting authors.}. We then explore the impact of input prompts with varying level of context on forecasting performance. The results indicate that incorporating news articles significantly improves performance, while using few-shot examples leads to a decline in accuracy. We find that larger models consistently outperform smaller models, highlighting the potential of LLMs in enhancing automated forecasting.

\end{abstract}

\section{Introduction}

Forecasting future event is an important discipline across various domains. It helps organizations and policymakers in strategic planning, ranging from health care, economics and finance to risk management~\cite{jin-etal-2021-forecastqa, suLargeLanguageModelsSLR2024}.
Humans rely on their judgment, domain expertise, and experience when forecasting events. However, human forecasting is limited in scalability. When faced with a large number of events to forecast or vast amounts of information to include for the forecasting oriented reasoning, predicting future manually is not easy. Automated systems \cite{regev2023futuretimelines,kawai2010chronoseeker,zouForecastingFutureWorld2022,jatowt2011Aextracting,naveedComprehensiveOverviewLarge2024} 
can speed up this forecasting process, especially as future-related information appears abundantly online in the form of speculations, plans, or expectations expressed by users \cite{jatowt2013multilingual} that can be then extracted using automatic means \cite{nakajima2020,Nakajima2018FutureRS,regev2023futuretimelines,jatowt2011Aextracting}. 
Although prior studies investigated if LLMs can be useful in future prediction, it is not clear how effective prompts should look like and what kind of contextual information helps the task most.

We explore in this paper how LLMs can  
facilitate the process of forecasting events when supplemented with contextual information.
LLMs are able to handle complex natural language tasks, 
and are often able to approach human-level performance in many areas~\cite{naveedComprehensiveOverviewLarge2024}. It is then interesting how well they perform in predicting the future - one of the most challenging, yet quite common, tasks that humans frequently perform.

We examine the forecasting capabilities of LLMs on binary events. We provide various inputs that progressively include more information related to the forecasting question and we measure the effect of such information on forecasting success. For this we first introduce a novel dataset of 614 binary forecasting questions about recent events, all with resolved outcomes. These questions are sourced from Metaculus\footnote{https://www.metaculus.com/questions/}, a forecasting platform where users can submit questions about future events. We collect relevant background information and generate summaries of these articles to provide concise and relevant context for the forecasting process of the LLMs.


\section{Related Datasets and Analyses}

\citet{zouForecastingFutureWorld2022} introduced Autocast, a dataset of 6,707 forecasting questions with a large corpus of corresponding news articles. The forecasting questions are obtained from various forecasting platforms, where users can submit and try to forecast questions about real events. The questions are resolved by the platform team, once sufficient evidence for their outcome is available. The authors used the questions of their dataset to evaluate different language models performance on forecasting. Their findings showed that the best evaluated model performs worse with 65\% accuracy than aggregate human forecasts with 92\% accuracy. 

ForecastQA~\cite{jin-etal-2021-forecastqa} by Jin et al. contains 10,392 forecasting questions, 5,903 of which are binary. Crowd workers created these questions based on news articles from the LexisNexis database. They assigned timestamps before the article's publication to each question, to ensure the outcomes are not known at this date. The actual outcomes were also included in the dataset. Bert-based models were tested on ForecastQA, with accuracy scores about 11\% lower than human forecasts.

Yuan et al.~\citet{yuanBackToFutureEXPTIME2023} explore temporal reasoning with LLMs, focusing on event forecasting and explanation generation. They introduced ExpTime, a dataset generated from a temporal knowledge graph suitable for fine-tuning models. Each entry includes a query, context, the correct answer, and reasoning explanation. The authors fine-tuned various Llama-based LLMs on the ExpTime dataset, showing substantial improvements in reasoning abilities.

The above datasets are however unsuitable for examining recently released LLMs in forecasting binary events, since their questions relate to events too far in the past. ForecastQA and ExpTime have no questions after 2019, and Autocast only includes questions until mid-2022. Additionally, the prior studies have not investigated how the forecasting capabilities of LLMs change with varying levels of context, such as progressively providing additional context and background information about the forecasting questions.

\section{Dataset Creation}

We first introduce a novel dataset of recent forecast questions and their outcomes to evaluate recently deployed LLMs. The forecasting questions in our dataset are collected from Metaculus, a collective prediction platform where users can create forecasting questions about real-world events and answer them. We filter for binary questions that have already been resolved.
Our dataset includes background information and the resolution criteria of each question, both provided by the question's creator on Metaculus. The resolution criteria define the condition under which a question resolves as yes or no.
Since we focus on testing recently deployed LLMs, all of the forecasting questions start after October 2021, which is after the training cutoff date of GPT-3.5-turbo, the model with the earliest training cutoff among those we tested.

We augment our dataset with news articles related to the forecasting questions, so that the LLMs can make an informed forecast. This imitates the process of human forecasting, where the latest available information is gathered before making an informed prediction \cite{jin-etal-2021-forecastqa, zouForecastingFutureWorld2022}.
In particular, we use Google News\footnote{https://news.google.com} to find news article related to the forecasting question. The news article must be published at least five days before forecasting questions's resolution date, to prevent information leakage. We choose this gap because questions are resolved by the Metaculus team when sufficient evidence is available, which may take some time. After obtaining the URLs to the news websites from Google News, we scrape the news article using the Python library Newspaper3k\footnote{https://github.com/codelucas/newspaper}. We only include forecasting questions for which we find at least three news articles. Due to the length of some news articles, we summarize them with GPT-3.5-turbo model. This step ensures that the provided input prompt remains concise, preventing the LLMs from being overwhelmed by too much context. 

Table \ref{tab:datasets_comparison} compares various properties of the different datasets mentioned in the related work with those of our dataset. 
Additionally, Appendix \ref{app:dataset} lists all the attributes of our dataset, as well as the distribution of the categories of forecasting questions. 

\begin{table*}[h!]
\small
    \centering
    \resizebox{\textwidth}{!}{
        \begin{tabular}{|c|c|c|c|c|c|c|c|c|}
            \toprule
            \textbf{Dataset} & \textbf{Types} & \textbf{Origin} & \textbf{Includes} &\textbf{From - To} & \textbf{T-Q} & \textbf{yes-Q} & \textbf{no-Q} & \textbf{BR-Q} \\
            
            \midrule
            Autocast & B, MC, N & \makecell{Metaculus, \\ GJO, \\ CSET Foretell} & \makecell{Q, B, \\ Publish Date, \\ Resolution Date, \\ NA Links} & \makecell{ Sep 1, 2015 \\- Jun 17, 2022} & 6,707 & 598 & 1,405 & 2,003 \\ 
            
            \midrule
            ForecastQA & B, MC & \makecell{Crowdsourced \\from news} & \makecell{Q, \\ Time Constraints} & \makecell{Dec 31, 2017\\ - Nov 21, 2019} & 10,392 & 2,693 & 2,626 & 5,319 \\
            
            \midrule
            ExpTime &  B, NS & \makecell{ Temporal \\ reasoning \\ KG} & \makecell{Q, B, \\ Rationale} & \makecell{Dec 1, 2005\\ - Oct 31, 2018} & 26,769 & 12,138 & 9,005 & 21,143 \\
            
            \midrule
            Our Dataset & B & \makecell{Metaculus} & \makecell{Q, B, R, \\ Publish Date, \\ Resolution Date, \\ NA Summary} & \makecell{Oct 1, 2021\\- Jun 15, 2024} & 614 & 223 & 391 & 614 \\
            
            \bottomrule
        \end{tabular}
    }
    \caption{Comparison of forecasting question datasets. The abbreviations are defined as follows: \textbf{T-Q} represents the total number of questions, \textbf{yes-Q} denotes the number of questions with a 'yes' ground truth, \textbf{no-Q} indicates the number of questions with a 'no' ground truth, and \textbf{BR-Q} refers to the number of binary resolved questions. The dataset types include: \textbf{B} for binary, \textbf{MC} for multiple-choice and \textbf{N} for numerical question, as well as \textbf{NS} for neutral samples (binary questions with ambiguous answers). In the 'Origin' column, \textbf{GJO} denotes Good Judgment Open and \textbf{KG} refers to Knowledge Graph, in the 'Includes' column \textbf{Q} stands for question, \textbf{B} for Background Information, \textbf{R} for resolution criteria and \textbf{NA} for news articles.}
    \label{tab:datasets_comparison}
\end{table*}

\section{Analysis}

We examine the forecasting capabilities of three LLMs of different release dates and parameter sizes. The oldest LLM we test is GPT-3.5-turbo, an optimized successor of the 175 billion parameter GPT-3 \cite{suLargeLanguageModelsSLR2024}. The training cutoff date of GPT-3.5-turbo was at September 2021\footnote{https://platform.openai.com/docs/models/gpt-3-5-turbo}. 
Alpaca-7B is a fine-tuned model based on Llama-7B and is the second LLM we test. It consists of 7 billion parameter and was released on March 14, 2023\footnote{https://crfm.stanford.edu/2023/03/13/alpaca.html}. 
We also examine Llama2-13B-chat, an upgraded successor to Llama with 13 billion parameters. This model incorporates more training data and is optimized for chat, having been trained until the end of July 2023 \cite{touvronHugo_Llama2_2023}.

To investigate the above LLMs, we provide five different input prompts with varying amounts of context. The first prompt consists only of the forecasting question, referred to as Q. In the second prompt, we include additional background information created by the question author on Metaculus, abbreviated as B. For the third prompt, we add the summaries of three related news articles, denoted as NA. The fourth prompt also incorporates the resolution criteria, abbreviated as R. In our final experiment, we supply few-shot examples consisting of two examples with questions, context news article and background information, along with the actual result to guide the LLM in forecasting. These few-shot examples are referred to as FS. Table \ref{tab:context_lengths} presents the average length of the context provided in the input prompts. Appendix \ref{app:prompts} shows also the different prompts in detail.

Due to the later training cutoff dates of Alpaca-7B and Llama2-13B-chat, we use a subset of our dataset to examine these LLMs. Specifically, we use only forecasting questions that were created after their training cutoff dates (249 questions for Alpaca-7B, 154 for Llama2-13B-chat).

\begin{table}[h!]
    \small
    \centering
    \begin{tabular}{|c|c|}
        \toprule
        \textbf{Context} & \textbf{Avg. Length (words)} \\
        \midrule
        Background Info. & 155 \\
        News article & 97 \\
        Resolution Criteria & 86 \\
        \bottomrule
    \end{tabular}
    \caption{Average length of context provided to the LLM in words.}
    \label{tab:context_lengths}
\end{table}

\section{Analysis Results}

Table \ref{tab:performance_metrics_results} shows 
the results of the forecasting experiments. GPT-3.5-turbo, the successor to the 175 billion parameter model GPT-3, achieves the highest success rates across all LLMs. Llama2-13B-chat demonstrates better results with its 13 billion parameters than the smaller model Alpaca-7B, which has 7 billion parameters.
The third prompt (which includes background information and news article) achieves the best score, comparable to the forecasts made with prompt four (with additional resolution criteria). Similar, for Alpaca-7B the third and fourth prompts yield the best results, with prompt four showing slightly better performance. Llama2-13B-chat achieves its best success rate with input prompt three. However, performance decreases significantly using prompt four. This behaviour may be attributed to the fact that Llama2-13B-chat forecasts only 154 questions due to its later training cutoff date. We observe a decreasing success rate across all three LLMs when adding few-shot examples to the prompt (prompt five). This decrease may result from the input prompts becoming quite large when two questions with additional context and their outcomes as examples are included, thus requiring the LLMs to handle a larger context. In Appendix \ref{app:analysis} we present confusion matrices for each experiment conducted with each LLM. These matrices enable a comprehensive analysis of the experimental results. 

\begin{figure}[h!]
    \centering
    \includegraphics[width=0.47\textwidth]{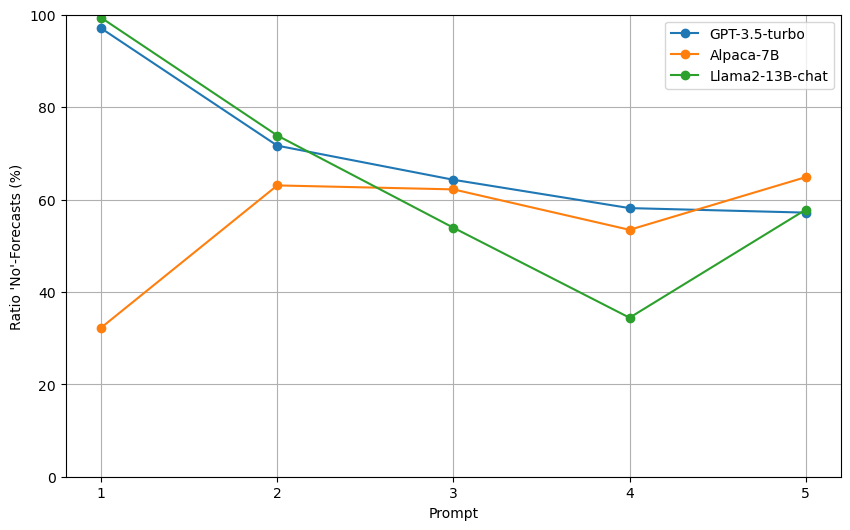}
    \caption{Ratio of questions forecasted as 'no' across forecasts made by various LLMs using different prompts. The numbers on Axis X has the following correspondence: 1: Q; 2: Q,B; 3: Q,B,NA; 4: Q,B,NA,R, 5: Q,B,NA,R,FS.}
    \label{fig:ratio_negative_forecasts}
\end{figure}

We observe in Figure \ref{fig:ratio_negative_forecasts} that when provided with only the question as input, GPT-3.5-turbo and Llama2-13B-chat forecast almost all questions as 'no'. As the context in the input prompt increases, the ratio of 'no' forecasts decreases for these two LLMs. However, forecasts made by Alpaca-7B show an increasing ratio of 'no' predictions with more context, eventually reaching a similar level as the other two LLMs.

Furthermore, we investigated the behaviour of the success rate concerning questions with increasing forecast horizon and question duration. The forecast horizon indicates the number of days a question is asked in the future from the training cutoff date of the LLM. Question duration refers to the time period in days during which a forecasting question was open for forecasts on Metaculus. The figures in Appendix \ref{app:evaluation}  provide a detailed illustration of how the success rate of forecasts varies across forecasting questions with increasing forecast horizon and question duration. Plots are shown for all three examined LLMs.

We found a slight trend of decreasing success rates for questions with increasing forecast horizons. No correlation was observed between success rates and question duration. We also asked the LLMs to provide a rationale for their forecasts and examined whether the length of the rationale correlates with the accuracy of the forecast. Our intention was to determine whether the LLMs generate longer rationales for incorrect forecasts, reflecting their uncertainty. However, we found no such correlation. 

\begin{table}[h!]
    \centering
    \resizebox{0.49\textwidth}{!}{
        \begin{tabular}{|c|c|c|c|c|c|c|}
            \toprule
            \textbf{\makecell{LLM \\ (Questions after)}} & \textbf{Prompt} & \textbf{\makecell{Success \\ Rate\footnotemark}} & \textbf{Precision} & \textbf{Recall} & \textbf{F1} \\
            \midrule
            \multirow{5}{*}{\makecell{GPT-3.5-turbo \\ (1 Oct. 2021)}} & Q & 0.64 & 0.50 & 0.04 & 0.07 \\
             & Q,B & 0.66 & 0.54 & 0.42 & 0.47 \\
             & Q,B,NA & \textbf{\underline{0.68}} & \underline{0.56} & 0.55 & 0.56 \\
             & Q,B,NA,R & \textbf{\underline{0.68}} & 0.55 & \underline{0.64} & \underline{0.59} \\
             & Q,B,NA,R,FS & 0.67 & 0.54 & 0.63 & 0.58 \\
            \midrule
            \multirow{5}{*}{\makecell{Alpaca-7B \\ (13 Mar. 2023)}} &   Q & 0.49 & 0.43 & \underline{0.70} & 0.54 \\
             & Q,B & 0.57 & 0.49 & 0.43 & 0.46 \\
             & Q,B,NA & 0.62 & \underline{0.55} & 0.50 & 0.52 \\
             & Q,B,NA,R & \underline{0.63} & \underline{0.55} & 0.61 & \underline{0.58} \\
             & Q,B,NA,R,FS & 0.59 & 0.53 & 0.44 & 0.48 \\
            \midrule
            \multirow{5}{*}{ \makecell{Llama2\\-13B-chat \\ (01 Aug. 2023)} } & Q & 0.57 & \textbf{\underline{1.00}} & 0.01 & 0.03 \\
             & Q,B & 0.62 & 0.60 & 0.36 & 0.45 \\
             & Q,B,NA  & \textbf{\underline{0.68}} & 0.62 & 0.66 & \textbf{\underline{0.64}} \\
             & Q,B,NA,R & 0.61 & 0.53 & \textbf{\underline{0.81}} & \textbf{\underline{0.64}} \\
             & Q,B,NA,R,FS & 0.57 & 0.51 & 0.49 & 0.50 \\
            \bottomrule
        \end{tabular}
    }
    \caption{Performance metrics for the examined LLMs across various prompts. The bold font indicates values best for a given metric and underline ones best for a given metric and LLM used.}
    \label{tab:performance_metrics_results}
\end{table}
\footnotetext{Success rate refers to accuracy.}

\section{Limitations}

Our study has a few limitations. Due to the recent training cutoff dates, we have 614 forecasting questions for GPT-3.5-turbo, 249 for Alpaca-7B and 154 for Llama2-13B-chat. The limited number of forecasting questions, particularly for the smaller LLMs, may lead to the presence of outliers. Furthermore, we obtained our forecasting questions from a single source, namely Metaculus. Additionally, the collected news articles were automatically searched via Google News and downloaded from their websites without manual analysis. As it sometimes happens with news genre, the news articles might also contain diverse kinds of biases \cite{farber2020multidimensional}.

\section{Conclusion}

In our study, we introduced a novel dataset of binary forecasting questions about real events submitted by users on the forecasting platform Metaculus. This dataset includes recent, already resolved forecasting question, making it suitable for examining the forecasting capabilities of recently deployed LLMs with a large parameter size. We augmented our dataset with at least three news article for each question and provide a concise, question related summary of them.
We examined three LLMs of varying model size and training dates, using prompts with different levels of context. Our findings indicate that these LLMs perform best when provided with context from news article, as well as background information and resolution criteria. However, excessively long input prompts containing few-shot examples negatively affect performance. Larger models outperform smaller models in forecasting binary future events.

However, the findings of this study could benefit from future research involving a greater number of forecasting questions and a wider variety of LLMs. Furthermore, exploring different prompts and prompt engineering methods could improve our results. In the future, we also plan to expand our dataset to incorporate sequences of future events, similar to \cite{regev2023futuretimelines} to extend the task to the construction of multiple future timelines \cite{yu-etal-2021-multi}.

\bibliography{main}

\appendix

\section{Dataset}
\label{app:dataset}

\subsection{Attributes}

Each instance of our dataset comprises the following properties:

\begin{itemize}
    \item \textbf{Question}: The binary forecasting question itself. 
    \item \textbf{Answer}: The actual outcome. 
    \item \textbf{Description}: Context or background information about the question, drafted by the questioner. It can also contain links to further information/websites.
    \item \textbf{Description text}: Consists only of the description text. Possible links to websites have been removed.
    \item \textbf{Resolution criteria}: The resolution criteria formulated by the question creator. It specifies the conditions under which the forecasting question will resolve as \highlight{yes} or \highlight{no}.
    \item \textbf{Resolution criteria text}: Consists only of the resolution criteria text. Possible links to websites have been removed.
    \item \textbf{ID}: The ID of the question on Metaculus. A unique number for each question.
    \item \textbf{Categories}: The categories of the topic about which the question is concerned.
    \item \textbf{GNews query}: Used for searching for news articles on Google News\footnote{https://news.google.com}. Derived from the question, shortened, and including only the most important information.
    \item \textbf{News article}: The news articles found with the support of Google News and crawled from the corresponding news website. Each news article has the following further attributes:
    \begin{itemize}
        \item \textbf{URL}: The URL to the news website from which the news article is crawled.
        \item \textbf{Title}: The title of the news article.
        \item \textbf{Text}: The full text of the news article crawled from the website.
        \item \textbf{Authors}: The authors of the news article.
        \item \textbf{Publish date}: The date when the news article was published.
        \item \textbf{Summary}: A summary of the news article consisting of a few sentences. It is created by the Python library Newspaper3k\footnote{https://github.com/codelucas/newspaper} used to crawl the article.
        \item \textbf{Summary LLM}: A summary of the news article, consisting of a few sentences which correspond directly to the question. It is created by GPT-3.5-turbo.
        \item \textbf{Keywords}: The most important keywords in the article, also created with the Python library Newspaper3k\footnote{https://github.com/codelucas/newspaper}.
    \end{itemize}
    \item \textbf{Created time}: The time when the forecasting question was created by the author on Metaculus.
    \item \textbf{Publish time}: Indicates the time when the question was released on Metaculus. This date is after the created time, as each question is revised by the admin team of Metaculus.
    \item \textbf{Resolve time}: Shows the date when the question is resolved.
    \item \textbf{Days open}: The number of days between the \textbf{Resolve time} and \textbf{Created time}.
\end{itemize}

\subsection{Categories}

\begin{figure}[H]
    \centering
    \includegraphics[width=0.47\textwidth]{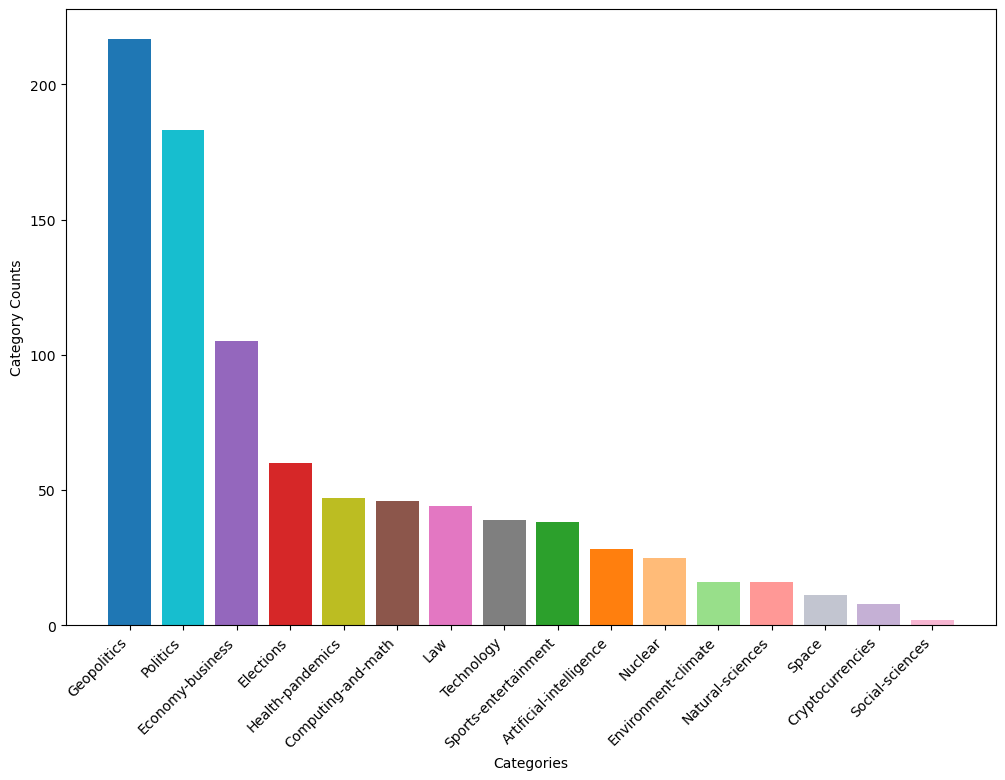}
    \caption{Distribution of categories.}
    \label{fig:datasets_cat}
\end{figure}

\section{Prompts}
\label{app:prompts}

We use the subsequent prompts to conduct the experiments with LLMs. In the first prompt, we provide only the forecasting question to the LLM, while the second prompt include additional background information. The third prompt incorporate three additional related news article, the fourth prompt the resolution criteria for the question, and the fifth prompt few extra examples for few-shot learning.

\subsection{Prompt 1: Only the Question}

\begin{lstlisting}[style=promptstyle, caption={The LLM receives only the forecasting question as input}]
system: 
"You are an assistant for forecasting events. You respond with 'yes' or 'no' to questions (delimited with <<<), depending on what you believe is correct and explain your decision (you provide a rationale). You delimit your  'yes' or 'no' with +++ (e.g.: +++yes+++). You delimit your rationale with --- (e.g.: ---rationale goes here). You attempt to make forecasts even when you don't know the answer from training."

user: 
"Your question, delimited with <<< (you must answer with 'yes' or 'no' (delimit it with +++) and provide then a rationale (start it with ---)):
<<<{question}<<<"
\end{lstlisting}

\subsection{Prompt 2: Adding Background Information}

\begin{lstlisting}[style=promptstyle, caption={Only background information is provided}]
system: 
"You are an assistant for forecasting events. You respond with 'yes' or 'no' to questions (delimited with <<<), depending on what you believe is correct and explain your decision (you provide a rationale). You delimit your  'yes' or 'no' with +++ (e.g.: +++yes+++). You delimit your rationale with --- (e.g.: ---rationale goes here). You consider the information provided under 'Background' delimited with #. You attempt to make forecasts even when you don't know the answer from training."

user: 
"Background: #{description}#
Your question, delimited with <<< (you must answer with 'yes' or 'no' (delimit it with +++) and provide then a rationale (start it with ---)):
<<<{question}<<<"
\end{lstlisting}

\subsection{Prompt 3: Adding News Articles}

\begin{lstlisting}[style=promptstyle, caption={Background information and three news article are provided}]
system: 
"You are an assistant for forecasting events. You respond with 'yes' or 'no' to questions (delimited with <<<), depending on what you believe is correct and explain your decision (you provide a rationale). You delimit your 'yes' or 'no' with +++ (e.g.: +++yes+++). You delimit your rationale with --- (e.g.: ---rationale goes here). You consider the information provided under 'Background', 'News Article 1', 'News Article 2' and 'News Article 3', each time delimited with #. You attempt to make forecasts even when you don't know the answer from training."

user: 
"Background: #{description}#
News Article 1: #{news1}#
News Article 2: #{news2}#
News Article 3: #{news3}#
Your question, delimited with <<< (you must answer with 'yes' or 'no' (delimit it with +++) and provide then a rationale (start it with ---)): 
<<<{question}<<< "
\end{lstlisting}

\subsection{Prompt 4: Adding Resolution Criteria}

\begin{lstlisting}[style=promptstyle, caption={Resolution criteria, background information and three news article are provided}]
system: "You are an assistant for forecasting events. You respond with 'yes' or 'no' to questions (delimited with <<<), depending on what you believe is correct and explain your decision (you provide a rationale). You delimit your  'yes' or 'no' with +++ (e.g.: +++yes+++). You delimit your rationale with --- (e.g.: ---rationale goes here). You consider the information provided under 'Resolution Criteria', 'Background', 'News Article 1', 'News Article 2' and 'News Article 3', each time delimited with #. You attempt to make forecasts even when you don't know the answer from training."

user:  
"Resolution Criteria: #{resolution_criteria}#
Background: #{description}#
News Article 1: #{news1}#
News Article 2: #{news2}#
News Article 3: #{news3}#
Your question, delimited with <<< (you must answer with 'yes' or 'no' (delimit it with +++) and provide then a rationale (start it with ---)):
<<<{question}<<<"
\end{lstlisting}

\subsection{Prompt 5: Adding Few-Shot Examples}

\begin{lstlisting}[style=promptstyle, caption={Resolution criteria, background information, three news article and few-shot examples are provided}]
system: "You are an assistant for forecasting events. You respond with 'yes' or 'no' to questions (delimited with <<<), depending on what you believe is correct and explain your decision (you provide a rationale). You delimit your  'yes' or 'no' with +++ (e.g.: +++yes+++). You delimit your rationale with --- (e.g.: ---rationale goes here). You consider the information provided under 'Resolution Criteria', 'Background', 'News Article 1', 'News Article 2' and 'News Article 3', each time delimited with #. You attempt to make forecasts even when you don't know the answer from training.

A few (solved) examples for you:

Question: 'Will the US have more than 6000 daily COVID-19 hospitalizations (7-day rolling average) before 1 January 2022?'
Background: 'The US has, in recent weeks, experienced a large uptick in COVID-19 cases (...)' 
Resolution Criteria: 'This question resolves positively if at any point (...)' 
Solution: 'yes' 

Question: 'Will the UK Conservative Party win the 2021 Batley and Spen by-election? ' 
Background: 'Batley and Spen is a constituency represented in the House of Commons (...)' 
Resolution Criteria: 'This question resolves positively if the Conservative (...)' 
Solution: 'no'"

user:  
"Resolution Criteria: #{resolution_criteria}#
Background: #{description}#
News Article 1: #{news1}#
News Article 2: #{news2}#
News Article 3: #{news3}#
Your question, delimited with <<< (you must answer with 'yes' or 'no' (delimit it with +++) and provide then a rationale (start it with ---)):
<<<{question}<<<"
\end{lstlisting}

\section{Analysis}
\label{app:analysis}

The following confusion matrices illustrate the results of our experiments across various prompts tested with the three different LLMs.

\subsection{Experiment 1: Only the Question}





\begin{figure}[H]
    \centering
    \begin{subfigure}[b]{0.23\textwidth}
        \centering
        \includegraphics[width=\textwidth]{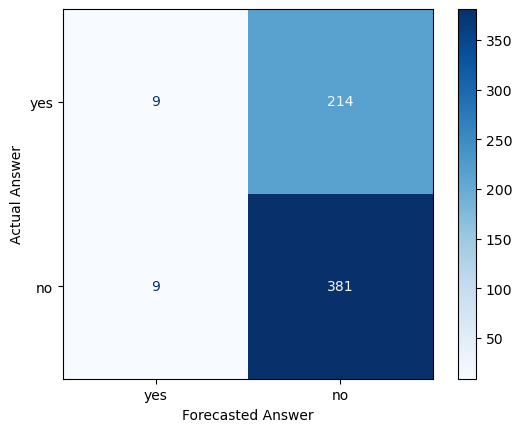}
        \caption{GPT-3.5-turbo}
        \label{confusionMatrix_GPT-3.5-turbo_1}
    \end{subfigure}
    \hfill
    \begin{subfigure}[b]{0.23\textwidth}
        \centering
        \includegraphics[width=\textwidth]{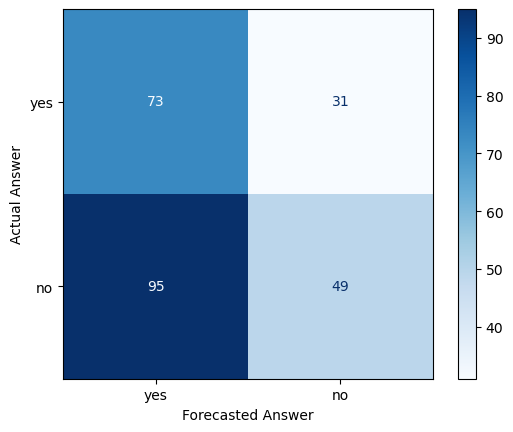}
        \caption{Alpaca-7B}
        \label{confusionMatrix_Alpaca-7B_1}
    \end{subfigure}
    \hfill
    \begin{subfigure}[b]{0.23\textwidth}
        \centering
        \includegraphics[width=\textwidth]{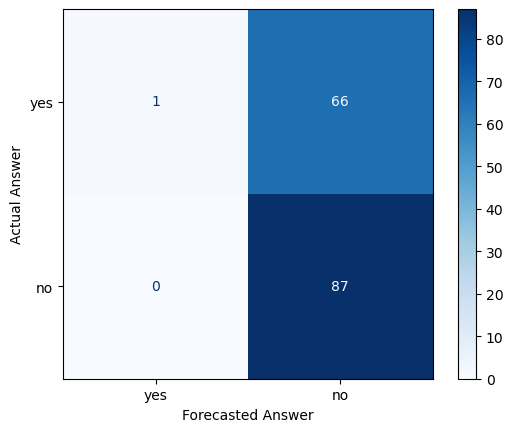}
        \caption{Llama2-13B}
        \label{confusionMatrix_Llama2-13B-chat_1}
    \end{subfigure}
    \caption{Confusion matrices for forecasts with only the question as input.}
    \label{fig:confusionMatrices_experiment1}
\end{figure}

\subsection{Experiment 2: Adding Background Information}

\begin{figure}[H]
    \centering
    \begin{subfigure}[b]{0.23\textwidth}
        \centering
        \includegraphics[width=\textwidth]{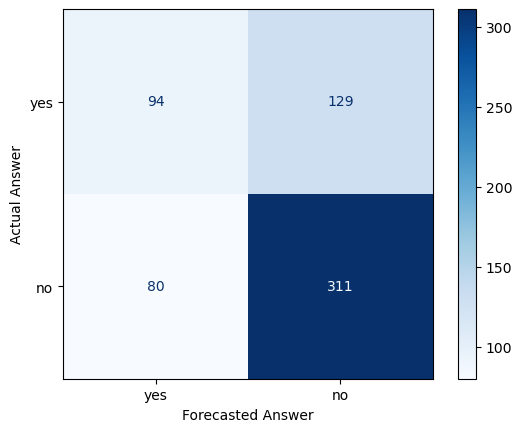}
        \caption{GPT-3.5-turbo.}
        \label{confusionMatrix_GPT-3.5-turbo_2}
    \end{subfigure}
    \hfill
    \begin{subfigure}[b]{0.23\textwidth}
        \centering
        \includegraphics[width=\textwidth]{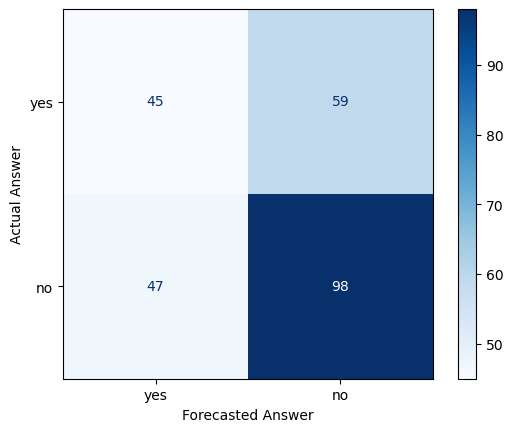}
        \caption{Alpaca-7B.}
        \label{confusionMatrix_Alpaca-7B_2}
    \end{subfigure}
    \hfill
    \begin{subfigure}[b]{0.23\textwidth}
        \centering
        \includegraphics[width=\textwidth]{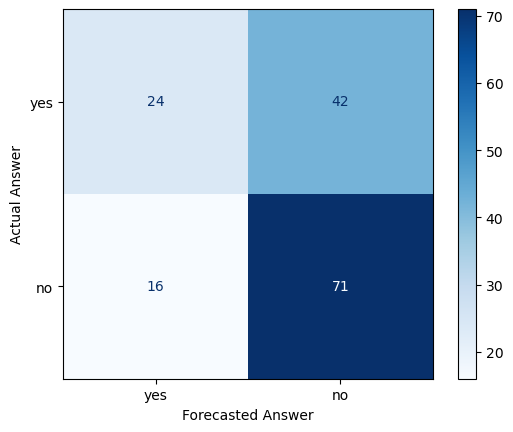}
        \caption{Llama2-13B-chat.}
        \label{confusionMatrix_Llama2-13B-chat_2}
    \end{subfigure}
    \caption{Confusion matrices for forecasts with question and background information as input.}
    \label{fig:confusionMatrices_experiment2}
\end{figure}

\subsection{Experiment 3: Adding News Articles}

\begin{figure}[H]
    \centering
    \begin{subfigure}[b]{0.23\textwidth}
        \centering
        \includegraphics[width=\textwidth]{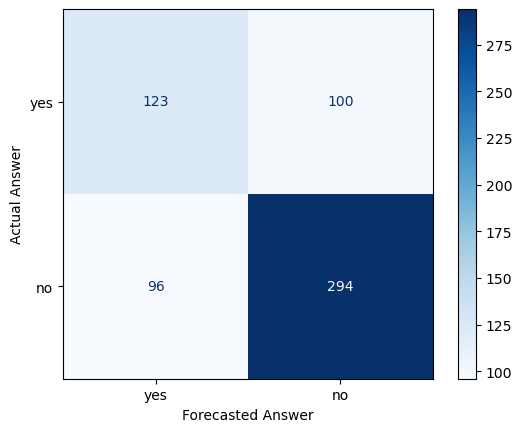}
        \caption{GPT-3.5-turbo.}
        \label{confusionMatrix_GPT-3.5-turbo_3}
    \end{subfigure}
    \hfill
    \begin{subfigure}[b]{0.23\textwidth}
        \centering
        \includegraphics[width=\textwidth]{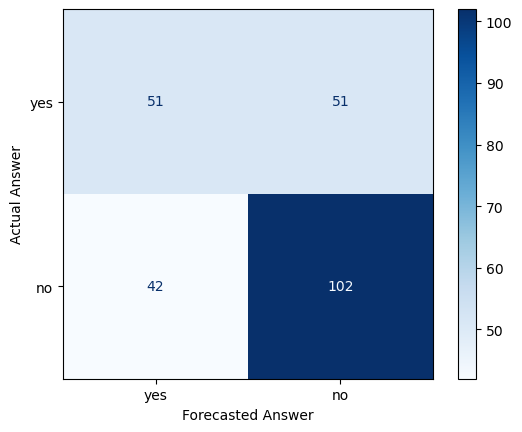}
        \caption{Alpaca-7B.}
        \label{confusionMatrix_Alpaca-7B_3}
    \end{subfigure}
    \hfill
    \begin{subfigure}[b]{0.23\textwidth}
        \centering
        \includegraphics[width=\textwidth]{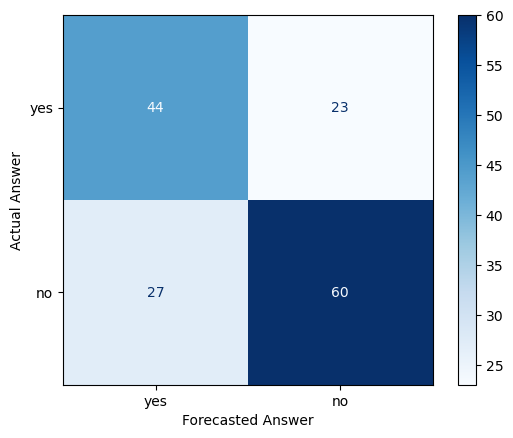}
        \caption{Llama2-13B-chat.}
        \label{confusionMatrix_Llama2-13B-chat_3}
    \end{subfigure}
    \caption{Confusion matrices for forecasts with question, background information, and news articles as input.}
    \label{fig:confusionMatrices_experiment3}
\end{figure}

\subsection{Experiment 4: Adding Resolution Criteria}

\begin{figure}[H]
    \centering
    \begin{subfigure}[b]{0.23\textwidth}
        \centering
        \includegraphics[width=\textwidth]{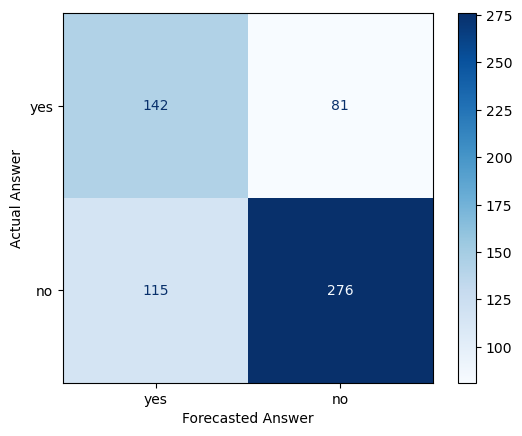}
        \caption{GPT-3.5-turbo.}
        \label{confusionMatrix_GPT-3.5-turbo_4}
    \end{subfigure}
    \hfill
    \begin{subfigure}[b]{0.23\textwidth}
        \centering
        \includegraphics[width=\textwidth]{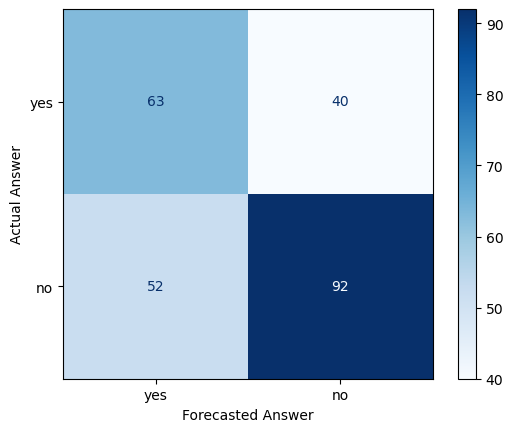}
        \caption{Alpaca-7B.}
        \label{confusionMatrix_Alpaca-7B_4}
    \end{subfigure}
    \hfill
    \begin{subfigure}[b]{0.23\textwidth}
        \centering
        \includegraphics[width=\textwidth]{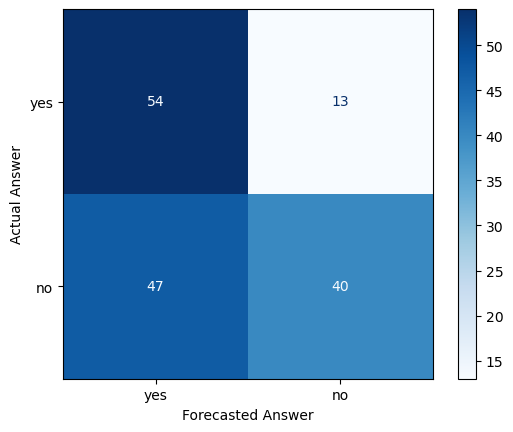}
        \caption{Llama2-13B-chat.}
        \label{confusionMatrix_Llama2-13B-chat_4}
    \end{subfigure}
    \caption{Confusion matrices for forecasts with question, background information, news articles, and resolution criteria as input.}
    \label{fig:confusionMatrices_experiment4}
\end{figure}

\subsection{Experiment 5: Adding Few-Shot Examples}

\begin{figure}[H]
    \centering
    \begin{subfigure}[b]{0.23\textwidth}
        \centering
        \includegraphics[width=\textwidth]{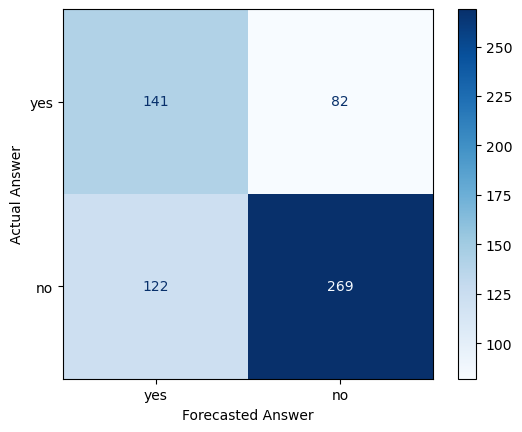}
        \caption{GPT-3.5-turbo.}
        \label{confusionMatrix_GPT-3.5-turbo_5}
    \end{subfigure}
    \hfill
    \begin{subfigure}[b]{0.23\textwidth}
        \centering
        \includegraphics[width=\textwidth]{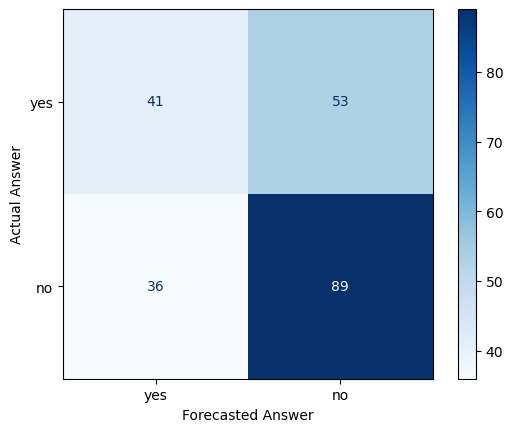}
        \caption{Alpaca-7B.}
        \label{confusionMatrix_Alpaca-7B_5}
    \end{subfigure}
    \hfill
    \begin{subfigure}[b]{0.23\textwidth}
        \centering
        \includegraphics[width=\textwidth]{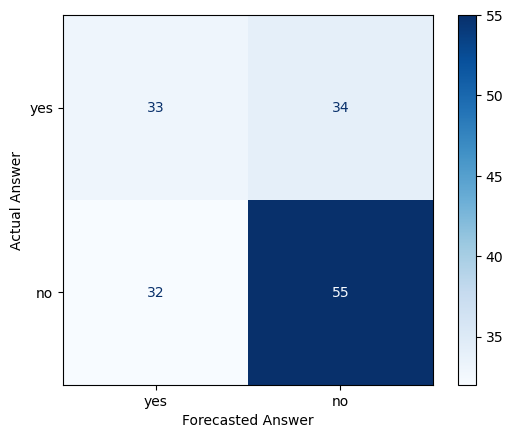}
        \caption{Llama2-13B-chat.}
        \label{confusionMatrix_Llama2-13B-chat_5}
    \end{subfigure}
    \caption{Confusion matrices for forecasts with question, background information, news articles, resolution criteria, and few-shot examples as input.}
    \label{fig:confusionMatrices_experiment5}
\end{figure}

\section{Evaluation}
\label{app:evaluation}

The following plots provide a detailed illustration of how the success rate of forecasts varies across forecasting questions with increasing forecast horizon and question duration. The plots are shown for all three examined LLMs.
The first sub-plot of each figure shows forecasts generated using only the question in the prompt, denoted as \textbf{Q}. The second plot presents forecasts that include background information, indicated by \textbf{B}. The third plot displays forecasts incorporating additional news articles, abbreviated as \textbf{NA}. The fourth plot illustrates forecasts based on prompts with added resolution criteria, denoted by \textbf{R}. The fifth plot presents forecasts that include few-shot examples in the prompt, abbreviated as \textbf{FS}.
We present figures showing the distribution of categories and number of questions across forecast horizon and question duration.

\subsection{Forecast Horizon}

\begin{figure}[H]
    \vskip\baselineskip
    \centering
    \begin{subfigure}[b]{0.23\textwidth} 
        \centering
          \includegraphics[width=\textwidth]{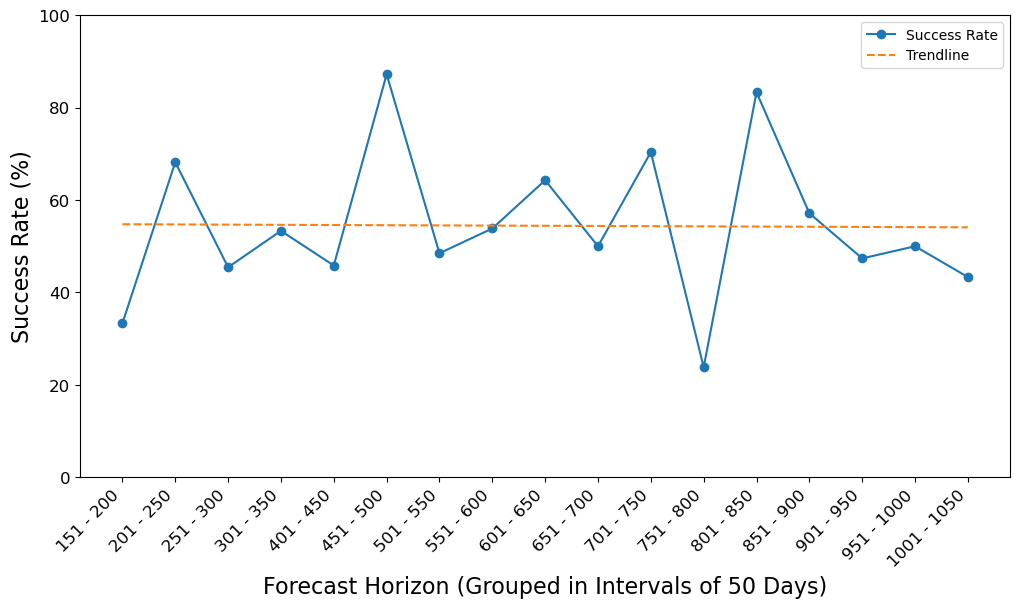}
        \caption{Input prompt: Q.}
        \label{fig:GPT-3.5-turbo_fh_1}
    \end{subfigure}
    \hfill
    \begin{subfigure}[b]{0.23\textwidth}
        \centering
        \includegraphics[width=\textwidth]{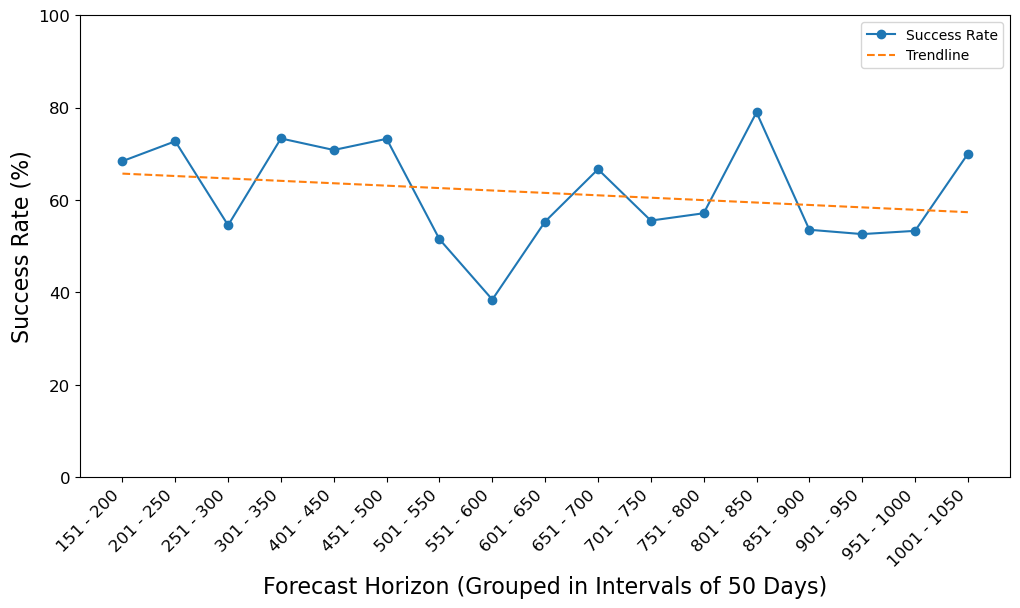}
        \caption{Input prompt: Q and B.}
        \label{fig:GPT-3.5-turbo_fh_2}
    \end{subfigure}
    
    \vskip\baselineskip

    \begin{subfigure}[b]{0.23\textwidth}
        \centering
        \includegraphics[width=\textwidth]{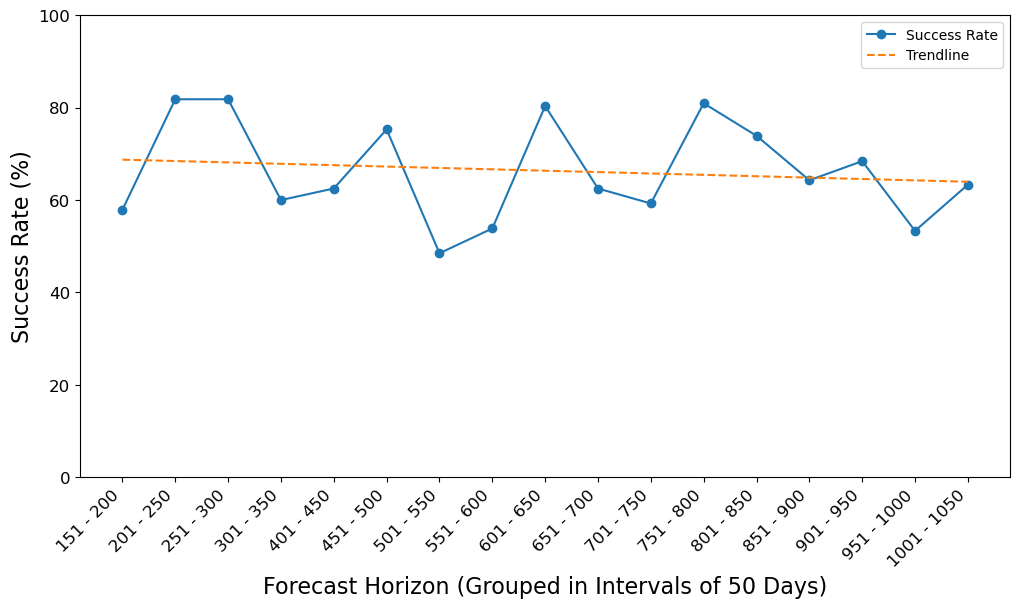}
        \caption{Input prompt: Q, B and NA.}
        \label{fig:GPT-3.5-turbo_fh_3}
    \end{subfigure}
    \hfill
    \begin{subfigure}[b]{0.23\textwidth}
        \centering
        \includegraphics[width=\textwidth]{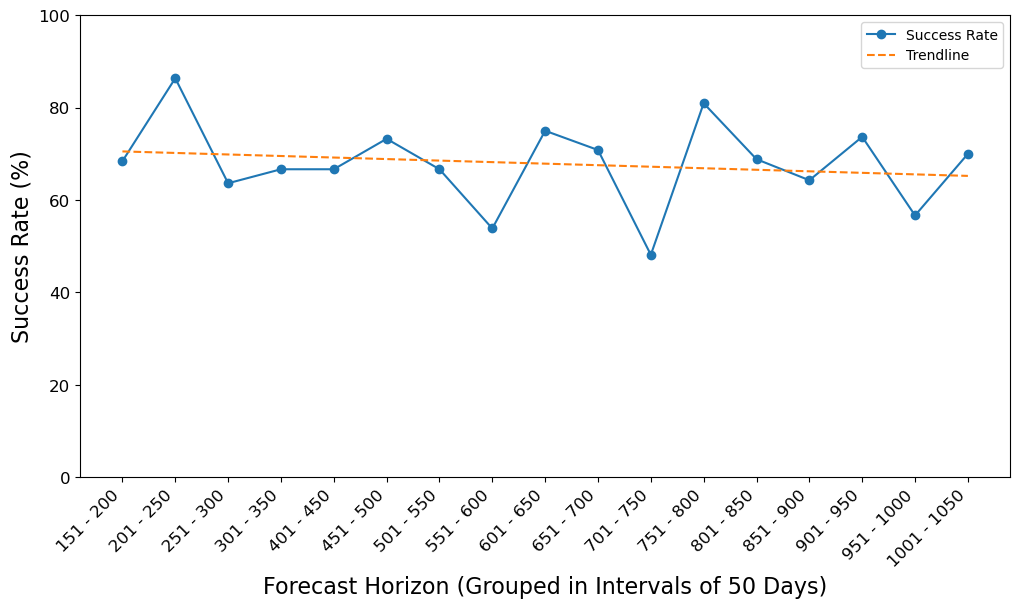}
        \caption{Input prompt: Q, B, NA and R.}
        \label{fig:GPT-3.5-turbo_fh_4}
    \end{subfigure}
    \vskip\baselineskip

    \begin{subfigure}[b]{0.23\textwidth}
        \centering
        \includegraphics[width=\textwidth]{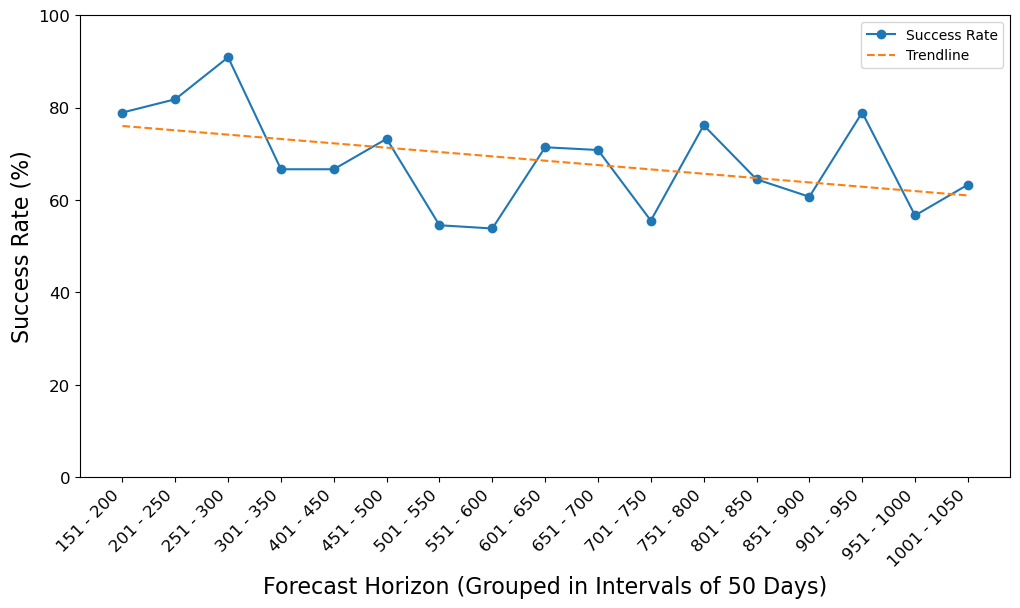}
        \caption{Input prompt: Q, B, NA, R and FS}
        \label{fig:GPT-3.5-turbo_fh_5}
    \end{subfigure}

    \caption{Success rate across forecast horizons of questions, grouped in 50-day intervals (from 151 to 1050 with a step of 50). Question forecasts were made by GPT-3.5-turbo using various input prompts.}
    \label{fig:GPT-3.5-turbo_forecast_horizon_plots}
\end{figure}


\begin{figure}[H]
    \raggedright
    \includegraphics[width=0.35\textwidth]{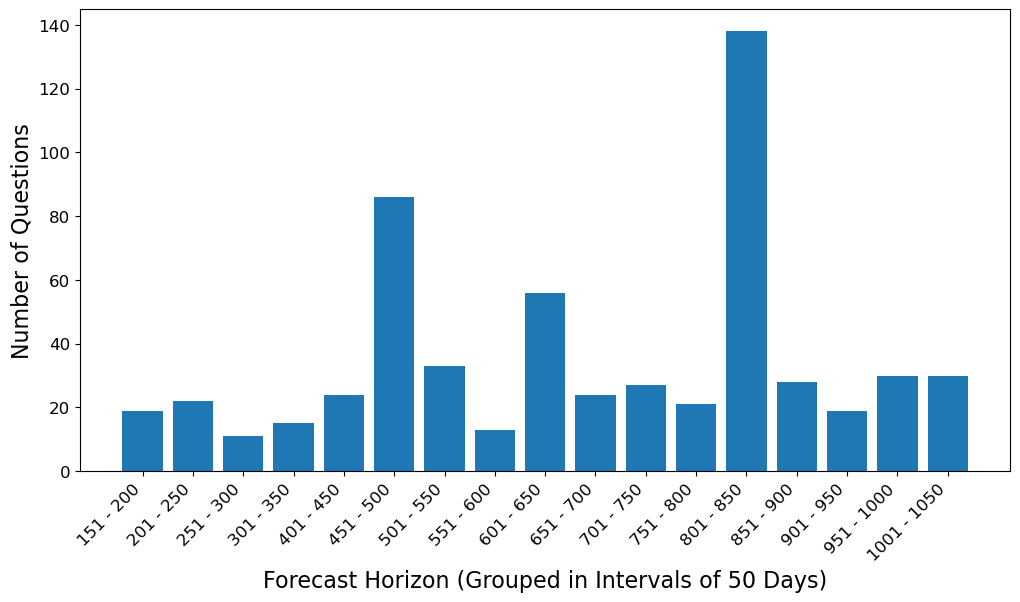}
    \caption{Question count across forecast horizon (GPT-3.5-turbo).}
    \label{fig:GPT-3.5_fh_orig_numQ}
\end{figure}

\begin{figure}[H]
    \raggedright
    \includegraphics[width=0.42\textwidth]{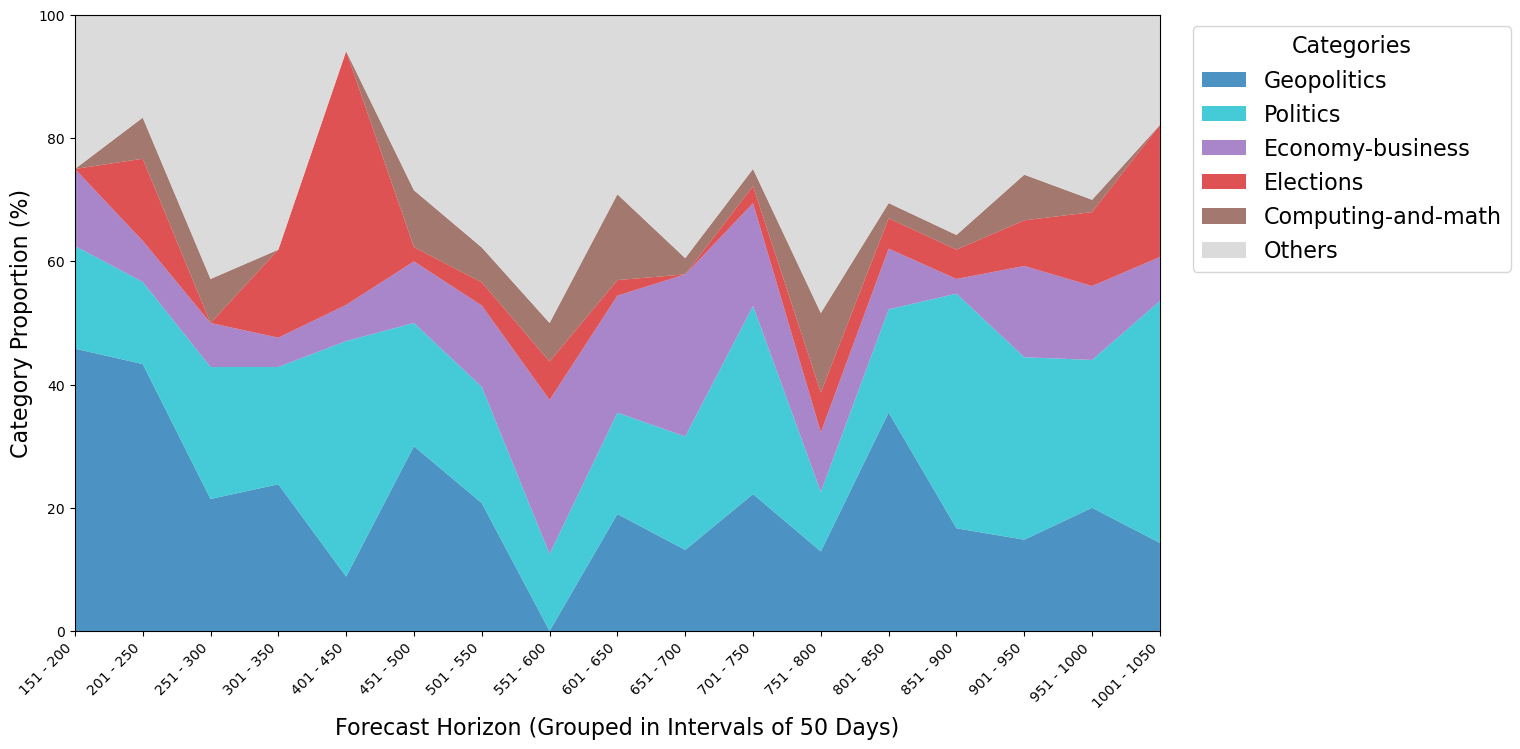}
    \caption{Ratio of the top five categories across forecast horizon (GPT-3.5-turbo).}
    \label{fig:GPT-3.5_fh_orig_cat}
\end{figure}

\begin{figure}[H]
    \vskip\baselineskip
    \centering
    \begin{subfigure}[b]{0.23\textwidth}
        \centering
        \includegraphics[width=\textwidth]{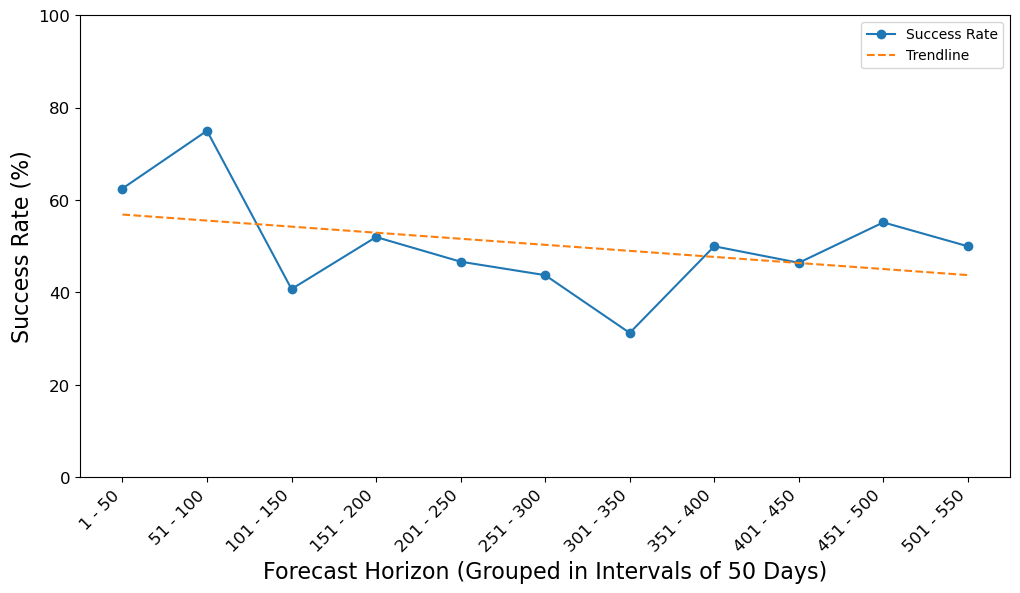}
        \caption{Input prompt: Q.}
        \label{fig:Alpaca-7B_fh_1}
    \end{subfigure}
    \hfill
    \begin{subfigure}[b]{0.23\textwidth}
        \centering
        \includegraphics[width=\textwidth]{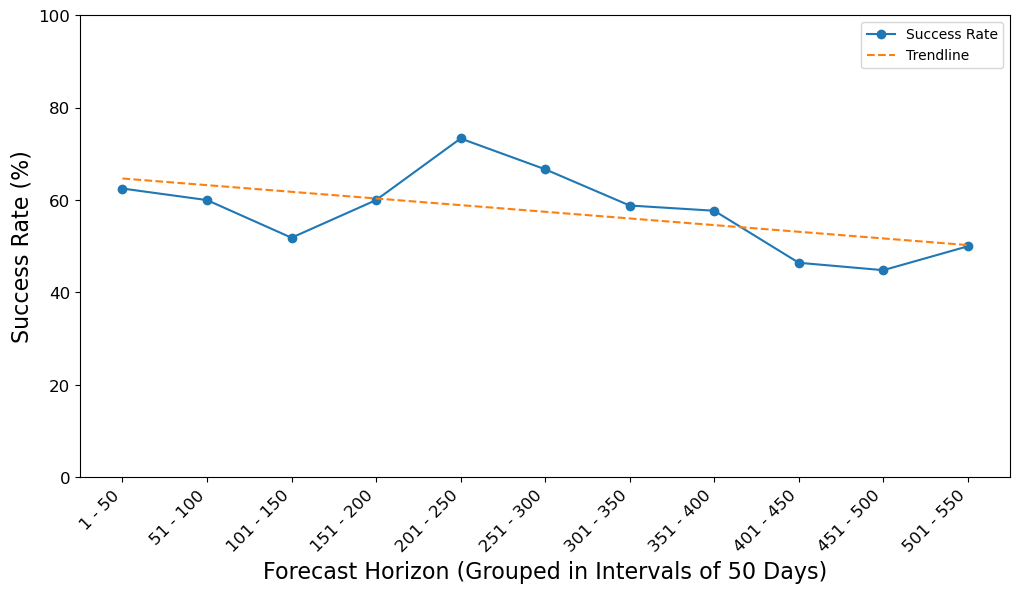}
        \caption{Input prompt: Q and B.}
        \label{fig:Alpaca-7B_fh_2}
    \end{subfigure}
    \vskip\baselineskip

    \begin{subfigure}[b]{0.23\textwidth}
        \centering
        \includegraphics[width=\textwidth]{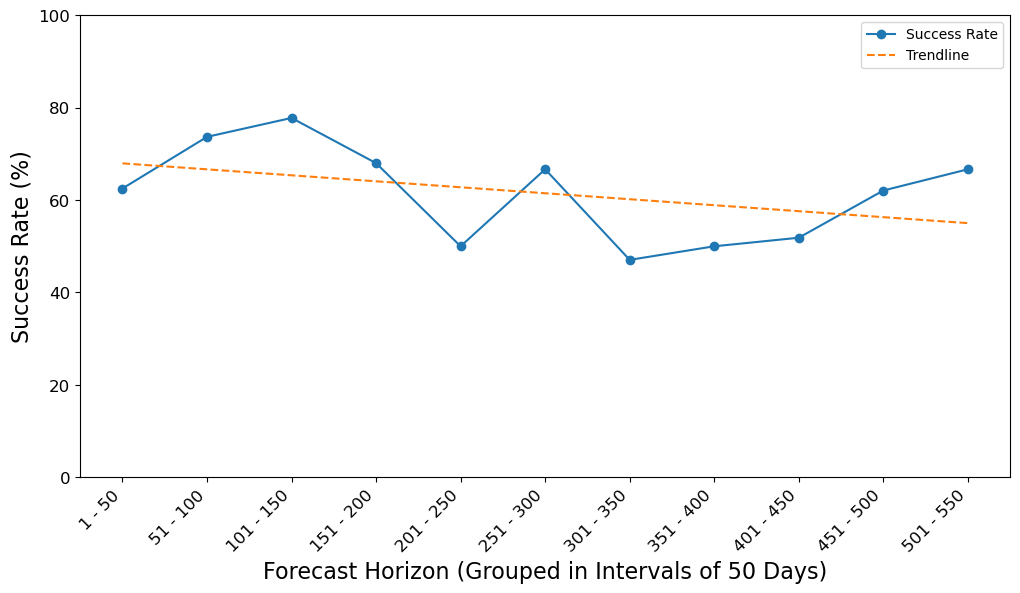}
        \caption{Input prompt: Q, B and NA.}
        \label{fig:Alpaca-7B_fh_3}
    \end{subfigure}
    \hfill
    \begin{subfigure}[b]{0.23\textwidth}
        \centering
        \includegraphics[width=\textwidth]{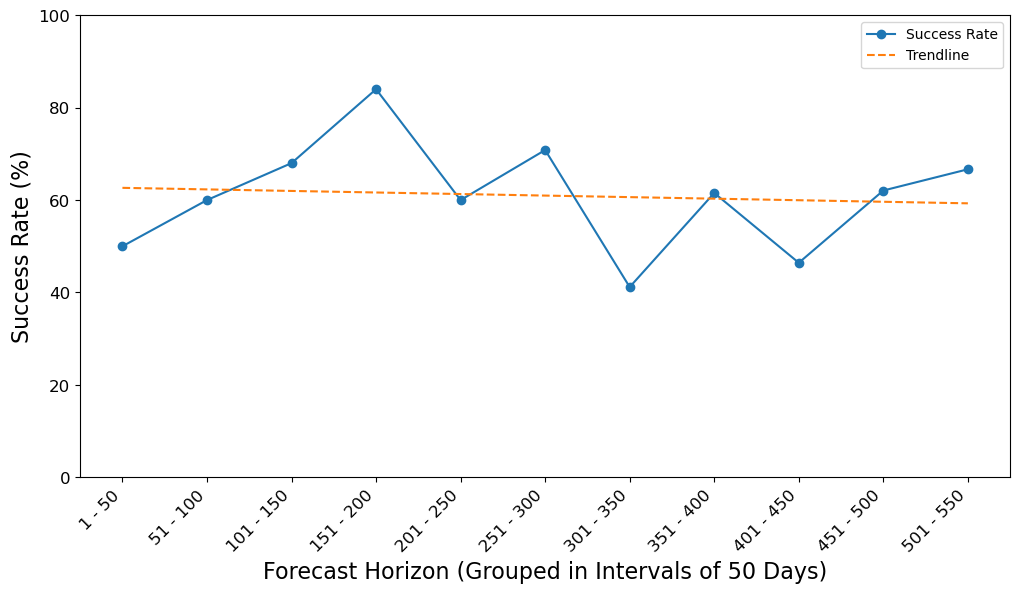}
        \caption{Input prompt: Q, B, NA and R.}
        \label{fig:Alpaca-7B_fh_4}
    \end{subfigure}
    \vskip\baselineskip

    \begin{subfigure}[b]{0.23\textwidth}
        \centering
        \includegraphics[width=\textwidth]{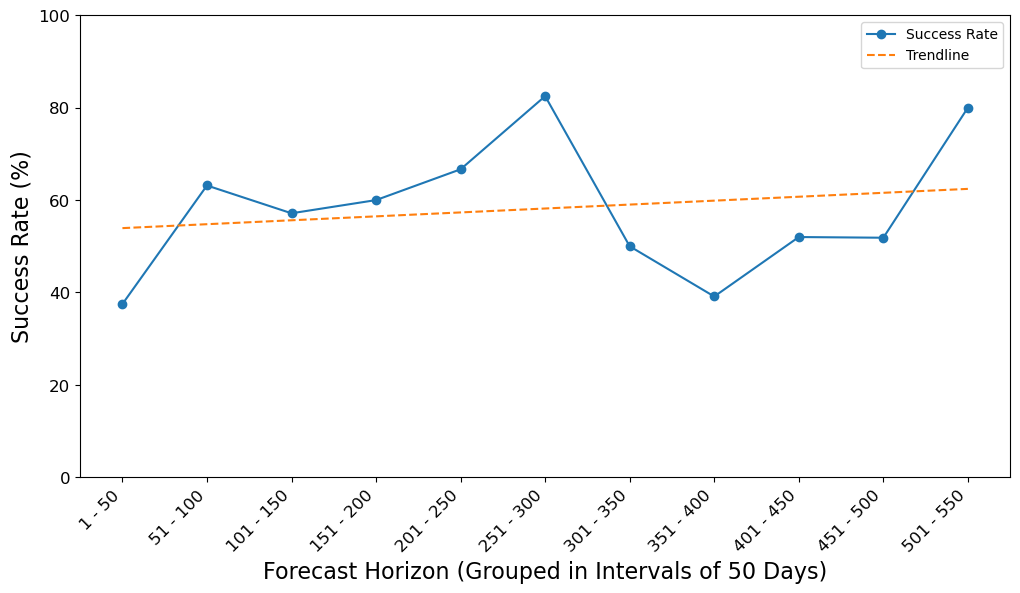}
        \caption{Input prompt: Q, B, NA, R and FS.} 
        \label{fig:Alpaca-7B_fh_5}
    \end{subfigure}
    
    \vskip\baselineskip
    \caption{Success rate across forecast horizons of questions, grouped in 50-day intervals (from 1 to 550 with a step of 50). Question forecasts were made by Alpaca-7B using various input prompts.}
    \label{fig:Alpaca-7B_forecast_horizon_plots}
\end{figure}


\begin{figure}[H]
    \raggedright
    \includegraphics[width=0.35\textwidth]{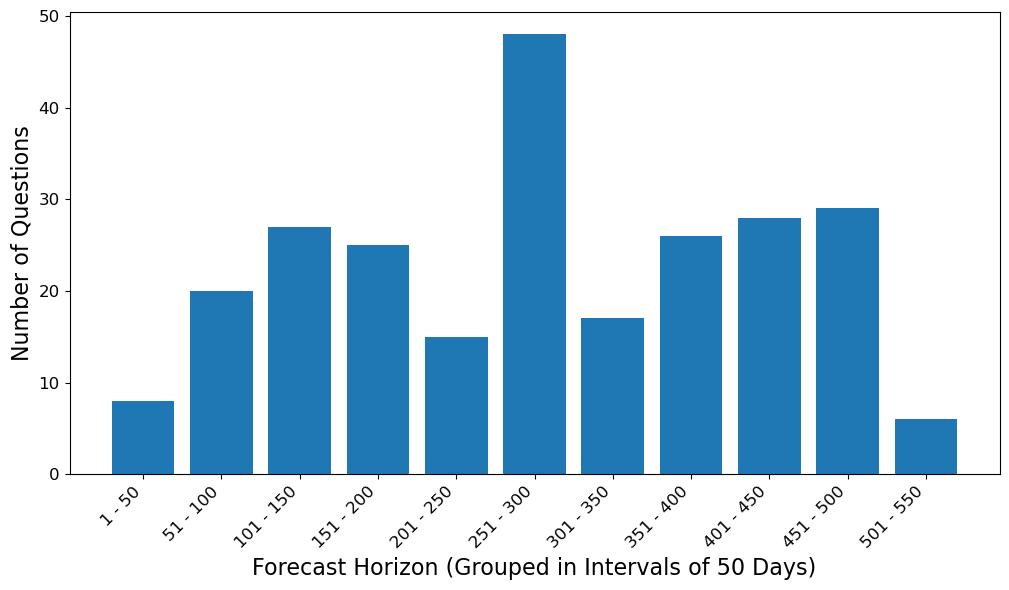}
    \caption{Question count across forecast horizon (Alpaca-7B).}
    \label{fig:Alpaca-7B_fh_orig_numQ}
\end{figure}

\begin{figure}[H]
    \raggedright
    \includegraphics[width=0.42\textwidth]{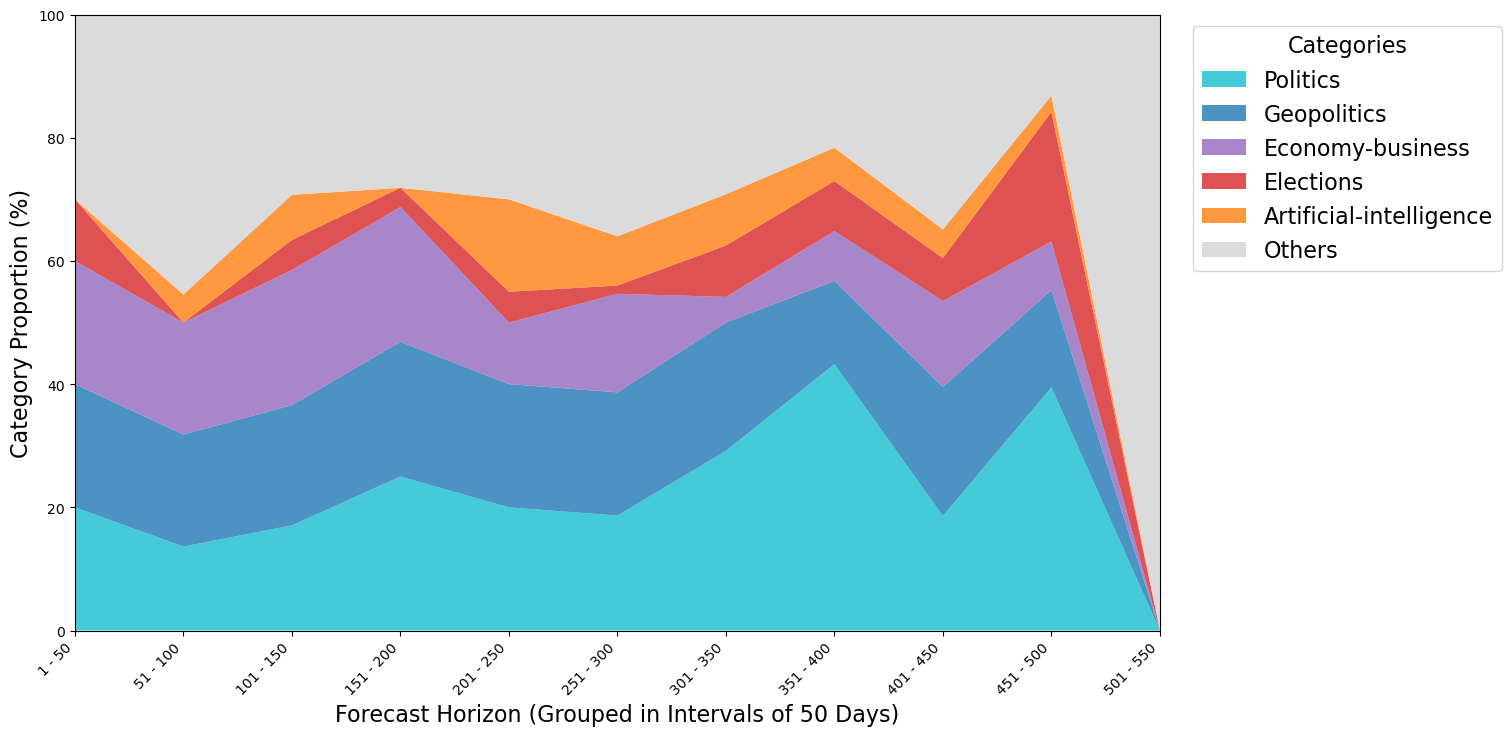}
    \caption{Ratio of the top five categories across forecast horizon (Alpaca-7B).}
    \label{fig:Alpaca-7B_fh_orig_cat}
\end{figure}

\begin{figure}[H]
    \vskip\baselineskip
    \centering
    \begin{subfigure}[b]{0.23\textwidth}
        \centering
        \includegraphics[width=\textwidth]{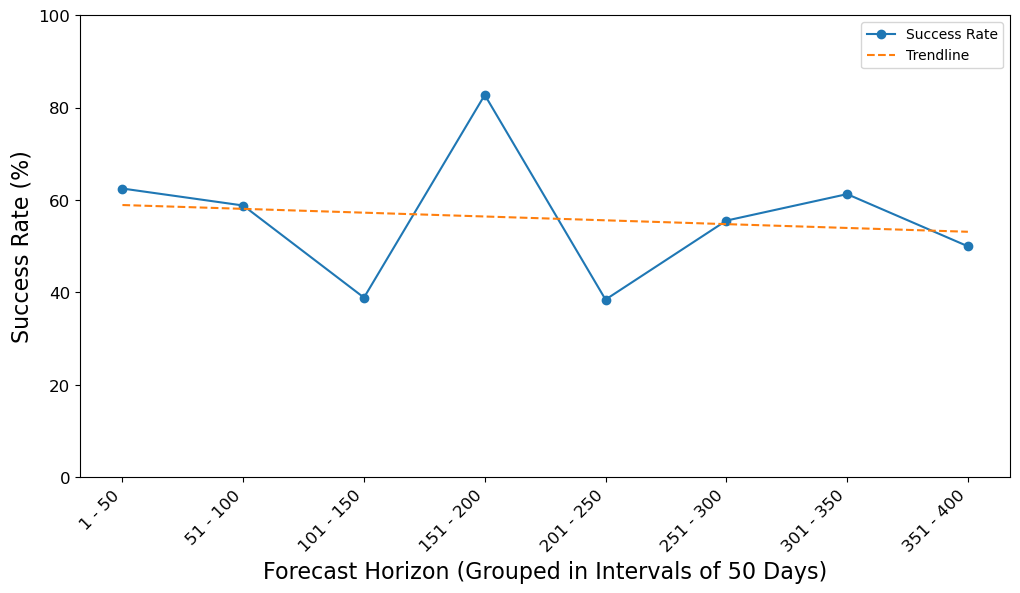}
        \caption{Input prompt: Q.}
        \label{fig:Llama2-13B_fh_1}
    \end{subfigure}
    \hfill
    \begin{subfigure}[b]{0.23\textwidth}
        \centering
        \includegraphics[width=\textwidth]{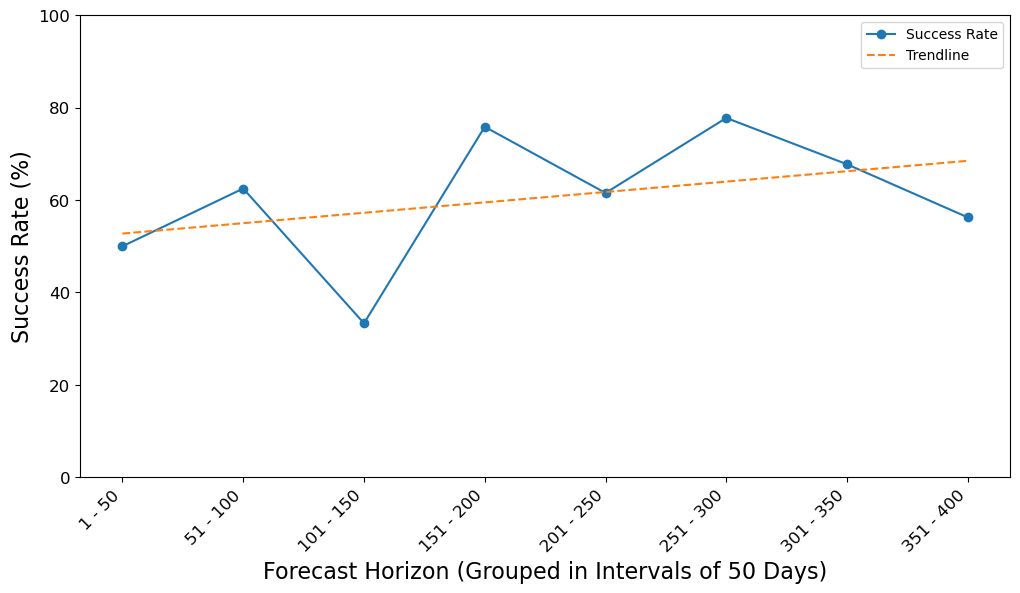}
        \caption{Input prompt: Q and B.}
        \label{fig:Llama2-13B_fh_2}
    \end{subfigure}

    \vskip\baselineskip

    \begin{subfigure}[b]{0.23\textwidth}
        \centering
        \includegraphics[width=\textwidth]{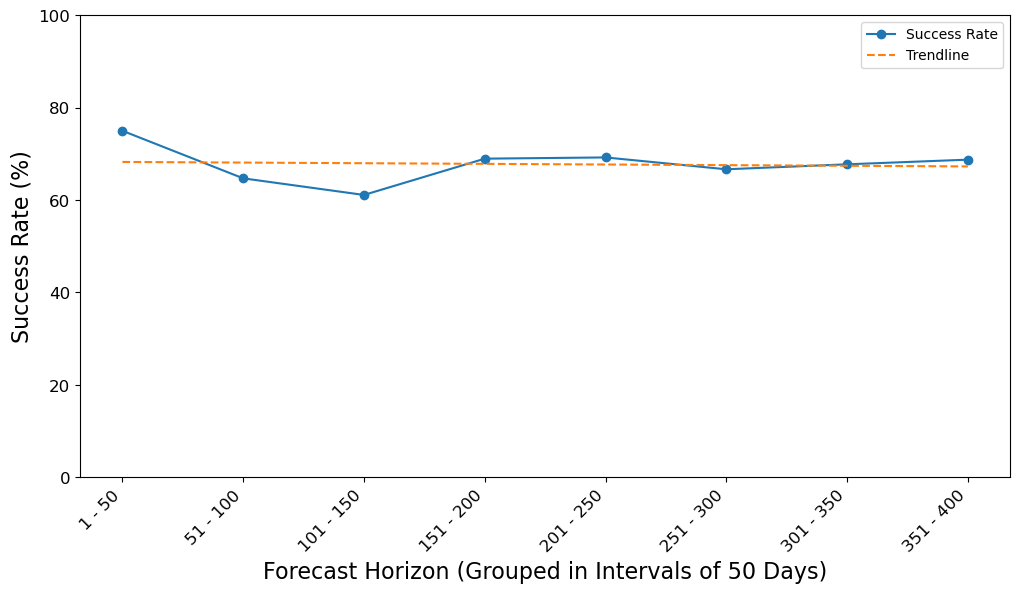}
        \caption{Input prompt: Q, B and NA.}
        \label{fig:Llama2-13B_fh_3}
    \end{subfigure}
    \hfill
    \begin{subfigure}[b]{0.23\textwidth}
        \centering
        \includegraphics[width=\textwidth]{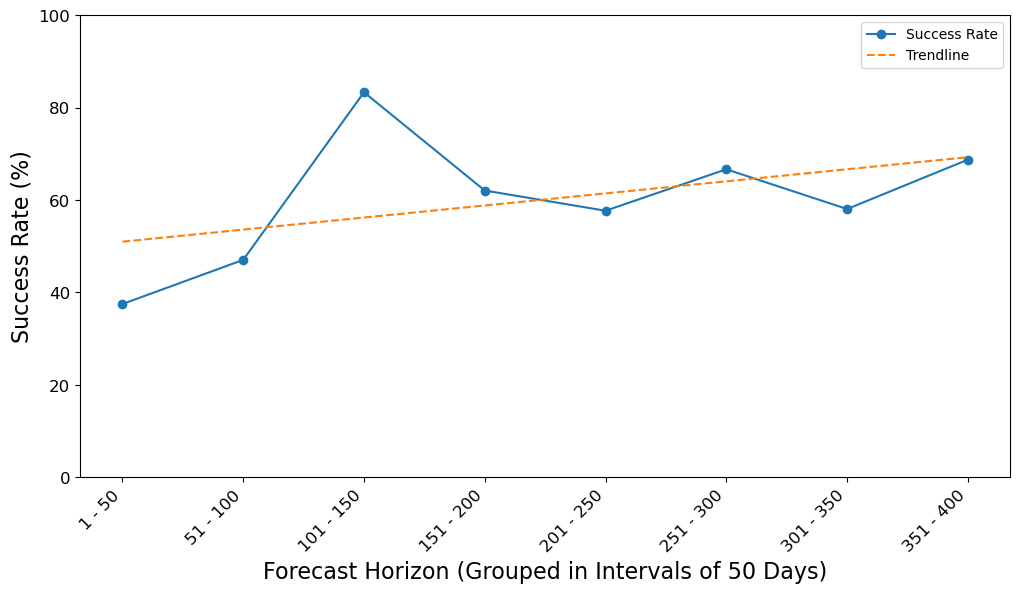}
        \caption{Input prompt: Q, B, NA and R.}
        \label{fig:Llama2-13B_fh_4}
    \end{subfigure}
    \vskip\baselineskip

    \begin{subfigure}[b]{0.23\textwidth}
        \centering
        \includegraphics[width=\textwidth]{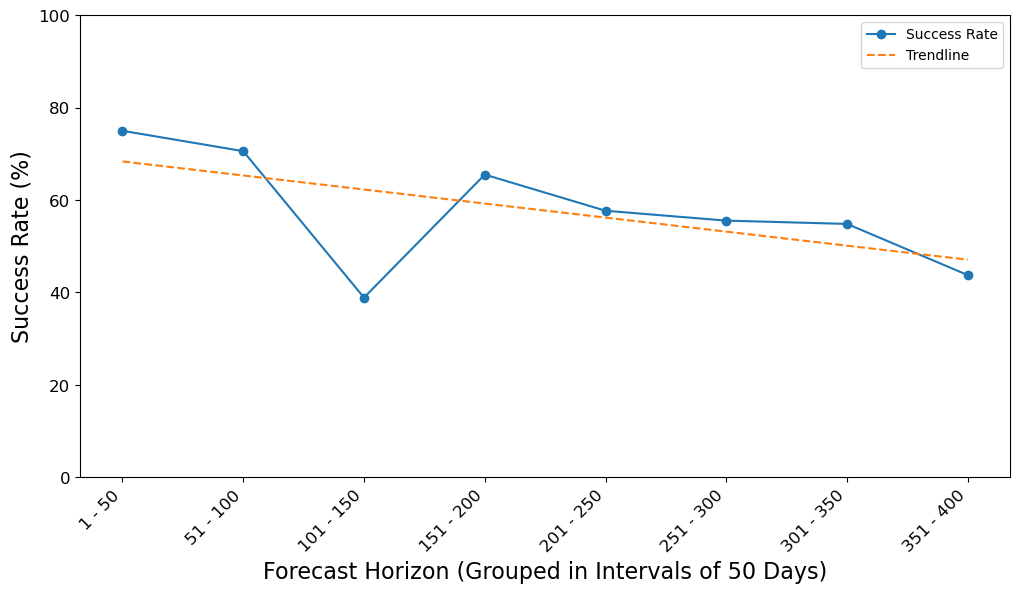}
        \caption{Input prompt: Q, B, NA, R and FS.}  
        \label{fig:Llama2-13B_fh_5}
    \end{subfigure}
    
    \vskip\baselineskip
    \caption{Success rate across forecast horizons of questions, grouped in 50-day intervals (from 1 to 400 with a step of 50). Question forecasts were made by Llama2-13B-chat using various input prompts.}
    \label{fig:Llama2-13B_forecast_horizon_plots}
\end{figure}

\begin{figure}[H]
    \raggedright
    \includegraphics[width=0.35\textwidth]{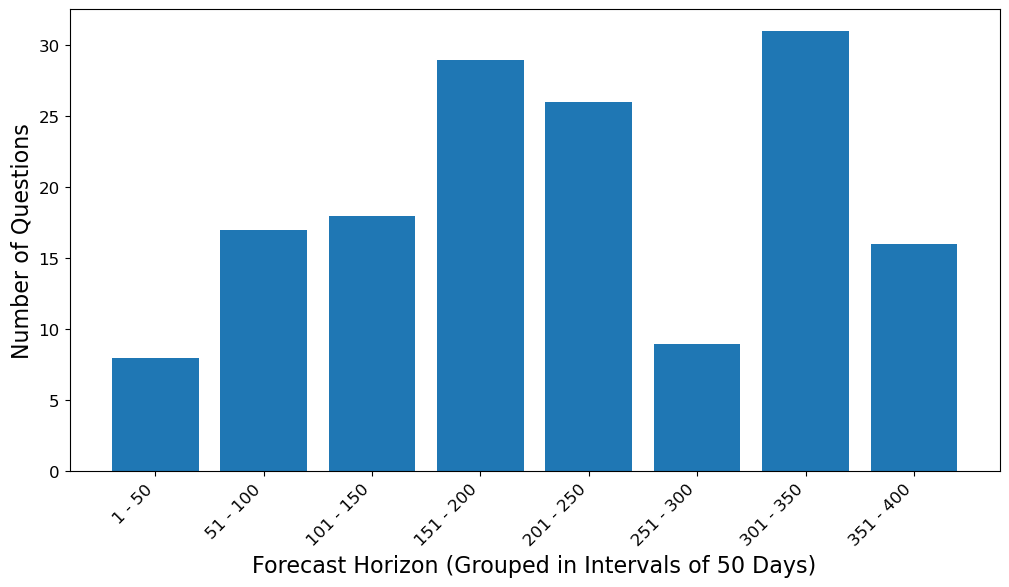}
    \caption{Question count across forecast horizon (Llama2-13B-chat).}
    \label{fig:Llama2-13B_fh_orig_numQ}
\end{figure}

\begin{figure}[H]
    \raggedright
    \includegraphics[width=0.42\textwidth]{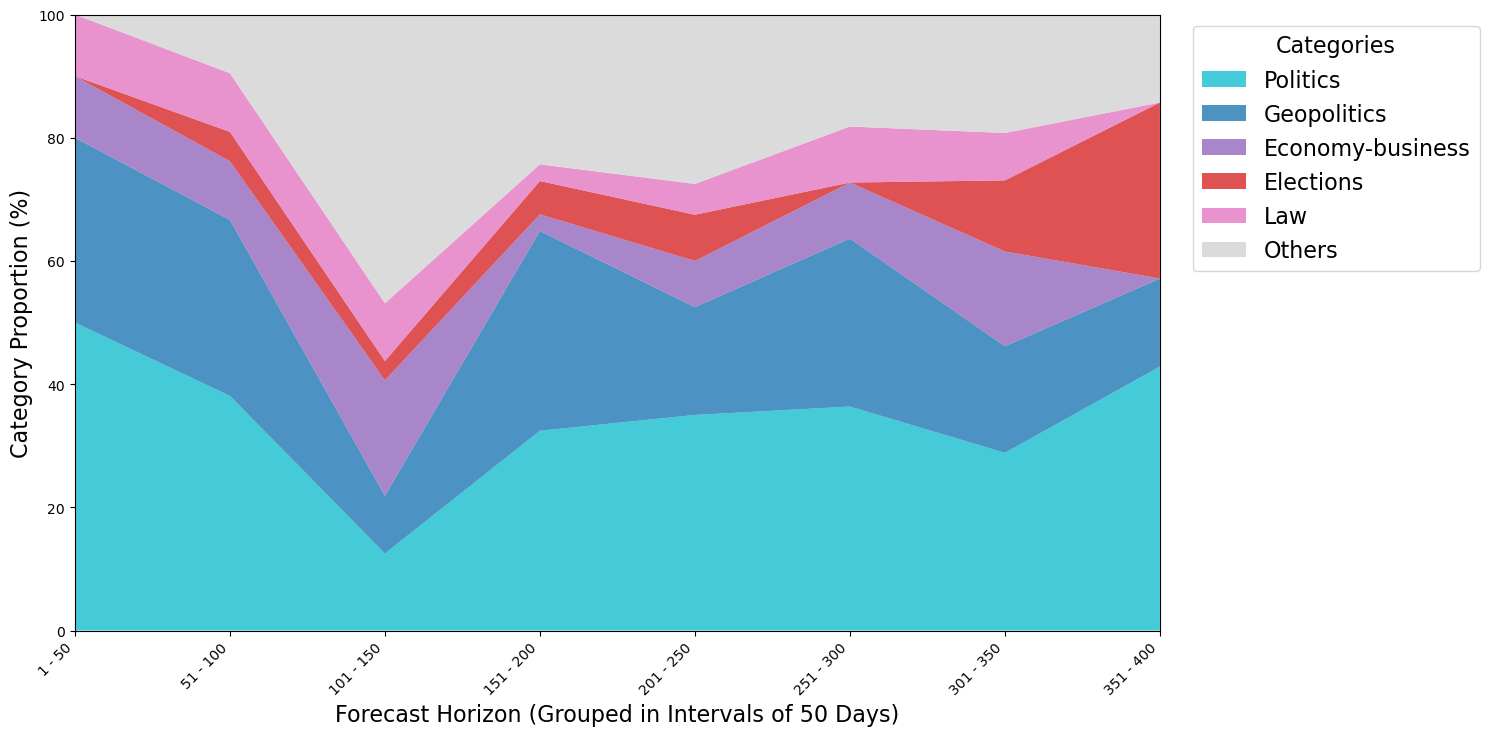}
    \caption{Ratio of the top five categories across forecast horizon (Llama2-13B-chat).}
    \label{fig:Llama2-13B_fh_orig_cat}
\end{figure}

\subsection{Question Duration}

\begin{figure}[H]
    \vskip\baselineskip
    \centering
    \begin{subfigure}[b]{0.23\textwidth}
        \centering
        \includegraphics[width=\textwidth]{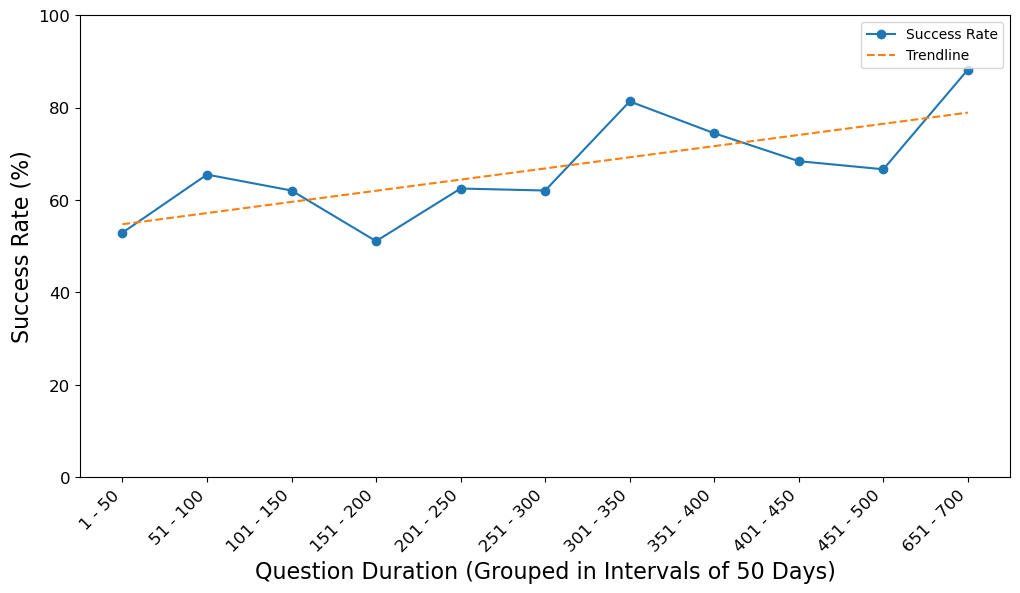}
        \caption{Input prompt: Q.}
        \label{fig:GPT-3.5-turbo_qd_1}
    \end{subfigure}
    \hfill
    \begin{subfigure}[b]{0.23\textwidth}
        \centering
        \includegraphics[width=\textwidth]{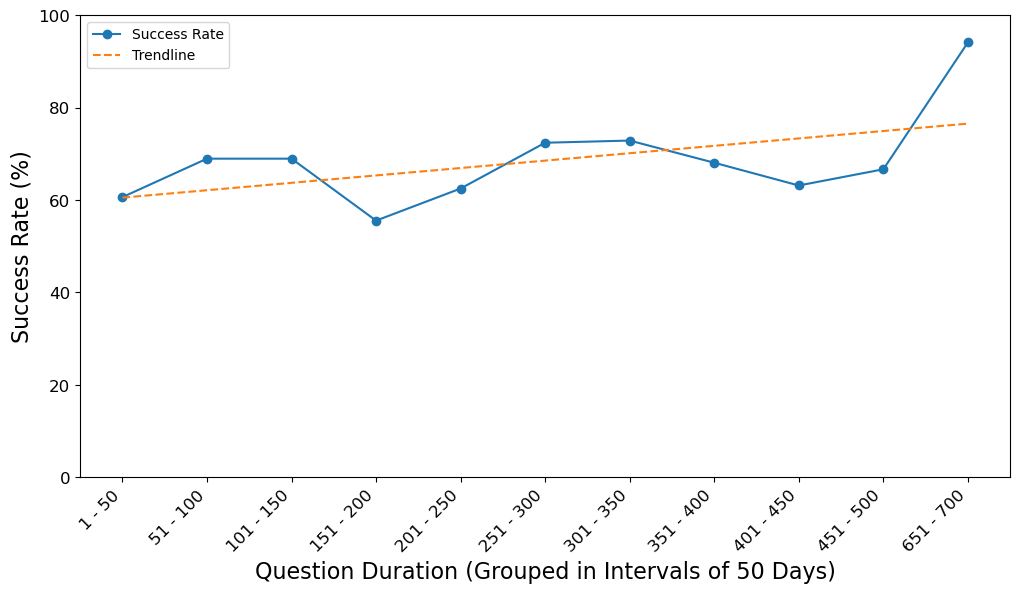}
        \caption{Input prompt: Q and B.}
        \label{fig:GPT-3.5-turbo_qd_2}
    \end{subfigure}
    \vskip\baselineskip

    \begin{subfigure}[b]{0.23\textwidth}
        \centering
        \includegraphics[width=\textwidth]{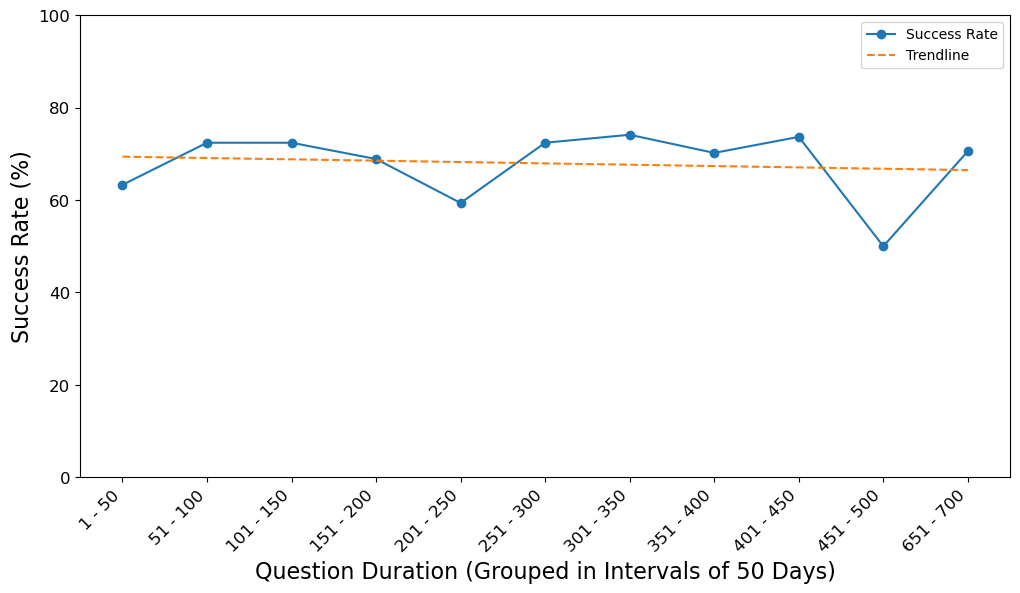}
        \caption{Input prompt: Q, B and NA.}
        \label{fig:GPT-3.5-turbo_qd_3}
    \end{subfigure}
    \hfill
    \begin{subfigure}[b]{0.23\textwidth}
        \centering
        \includegraphics[width=\textwidth]{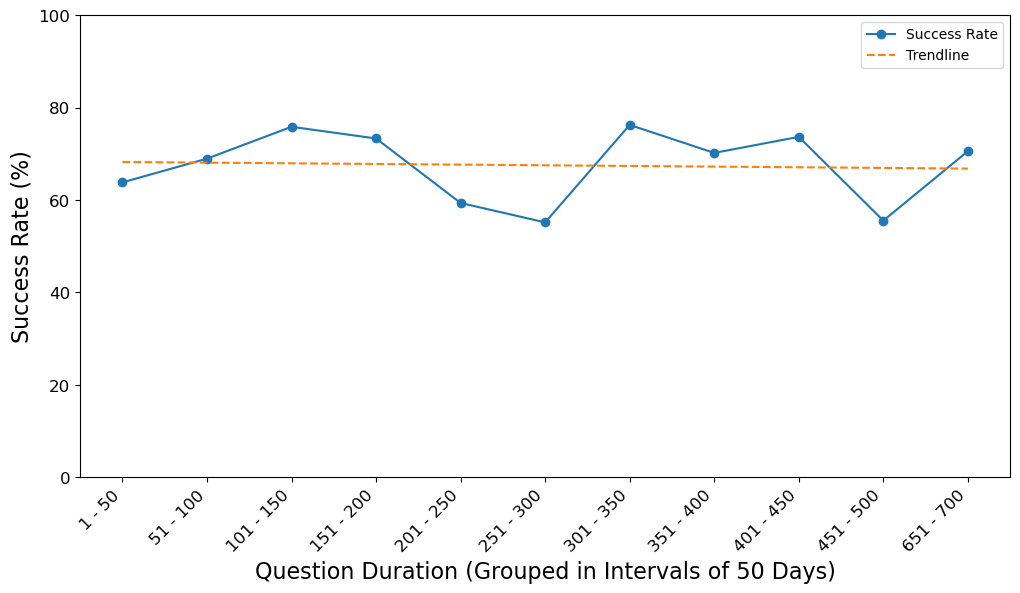}
        \caption{Input prompt: Q, B, NA and R.}
        \label{fig:GPT-3.5-turbo_qd_4}
    \end{subfigure}
    \vskip\baselineskip

    \begin{subfigure}[b]{0.23\textwidth}
        \centering
        \includegraphics[width=\textwidth]{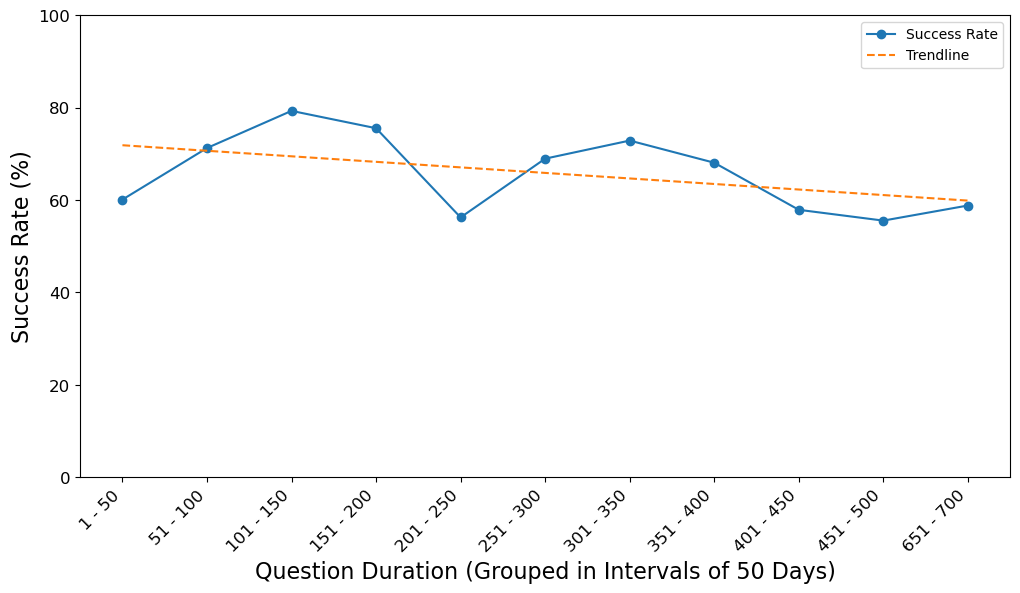}
        \caption{Input prompt: Q, B, NA, R and FS.}  
        \label{fig:GPT-3.5-turbo_qd_5}
    \end{subfigure}
    \vskip\baselineskip

    \caption{Success rate across question duration, grouped in 50-day intervals (from 1 to 700 with a step of 50). Question forecasts were made by GPT-3.5-turbo using various input prompts.}
    \label{fig:GPT-3.5-turbo_question_duration_plots}
\end{figure}

\begin{figure}[H]
    \raggedright
    \includegraphics[width=0.35\textwidth]{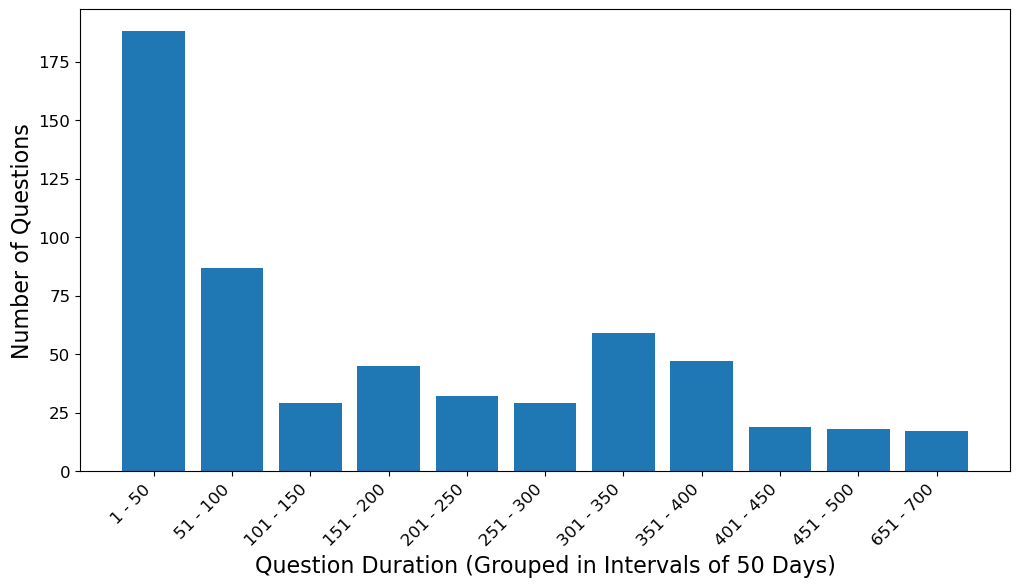}
    \caption{Question count across question duration (GPT-3.5-turbo).}
    \label{fig:GPT-3.5_qd_orig_numQ}
\end{figure}

\begin{figure}[H]
    \raggedright
    \includegraphics[width=0.42\textwidth]{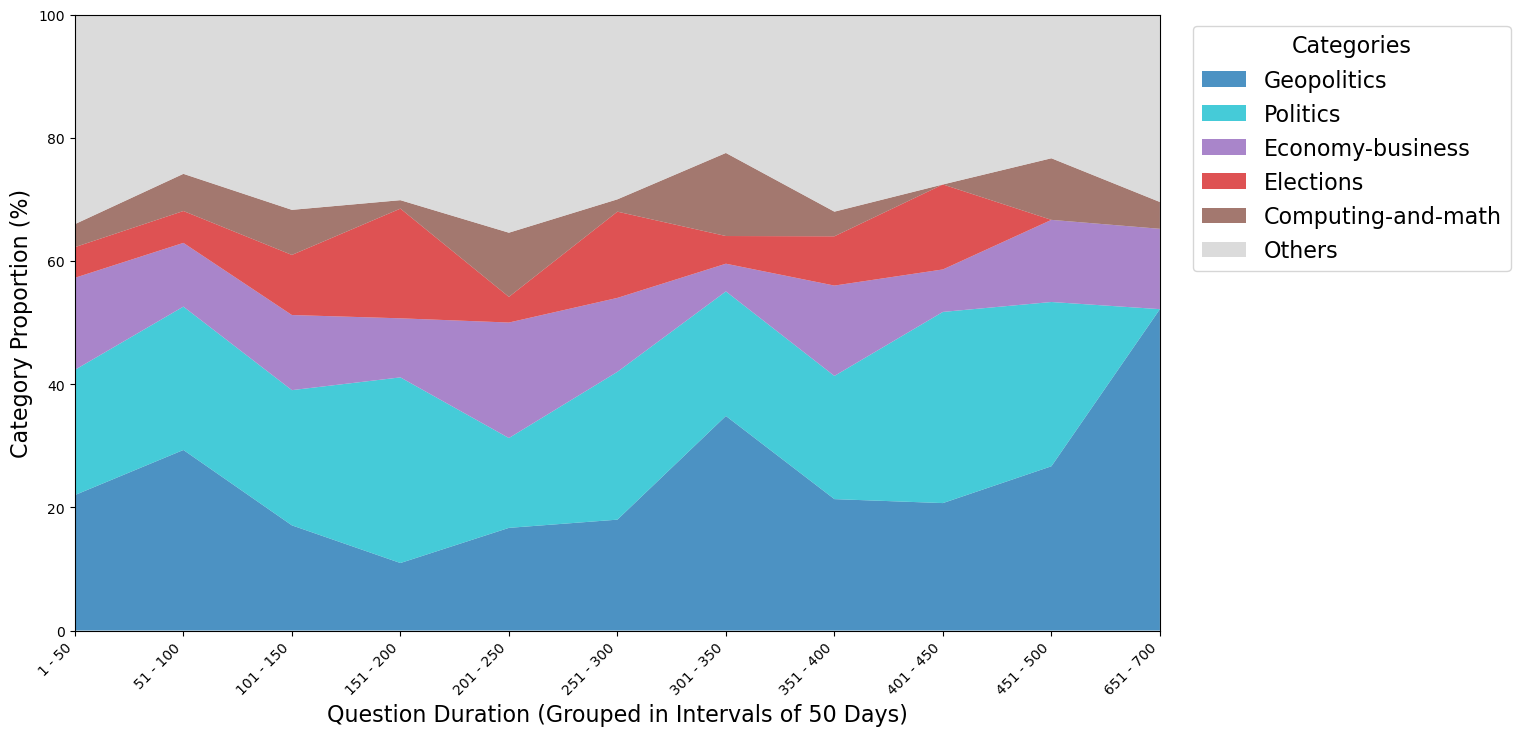}
    \caption{Ratio of the top five categories across question duration (GPT-3.5-turbo).}
    \label{fig:GPT-3.5_qd_orig_cat}
\end{figure}

\begin{figure}[H]
    \vskip\baselineskip
    \centering
    \begin{subfigure}[b]{0.23\textwidth}
        \centering
        \includegraphics[width=\textwidth]{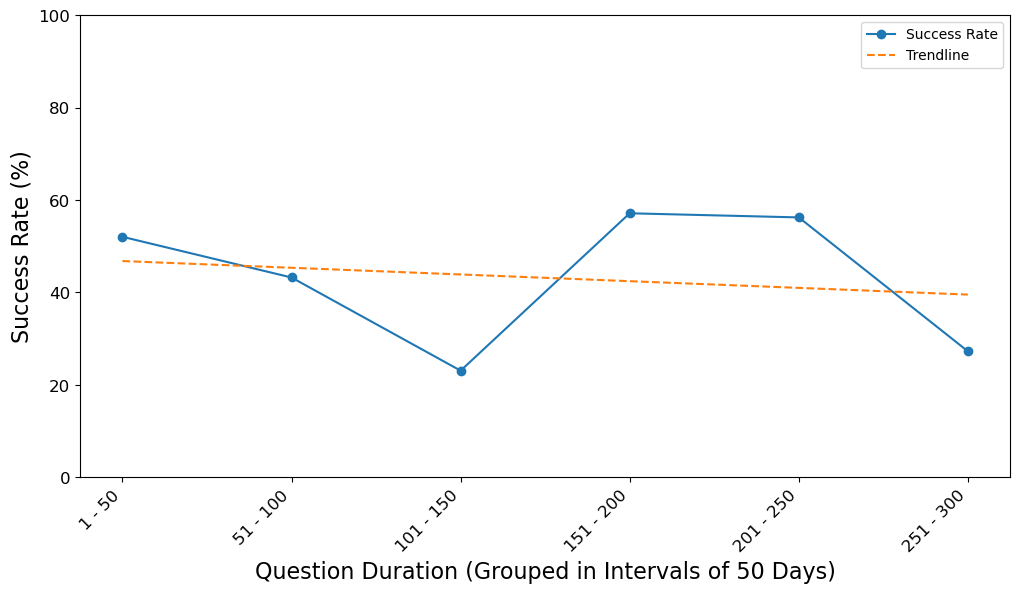}
        \caption{Input prompt: Q.}
        \label{fig:Alpaca-7B_qd_1}
    \end{subfigure}
    \hfill
    \begin{subfigure}[b]{0.23\textwidth}
        \centering
        \includegraphics[width=\textwidth]{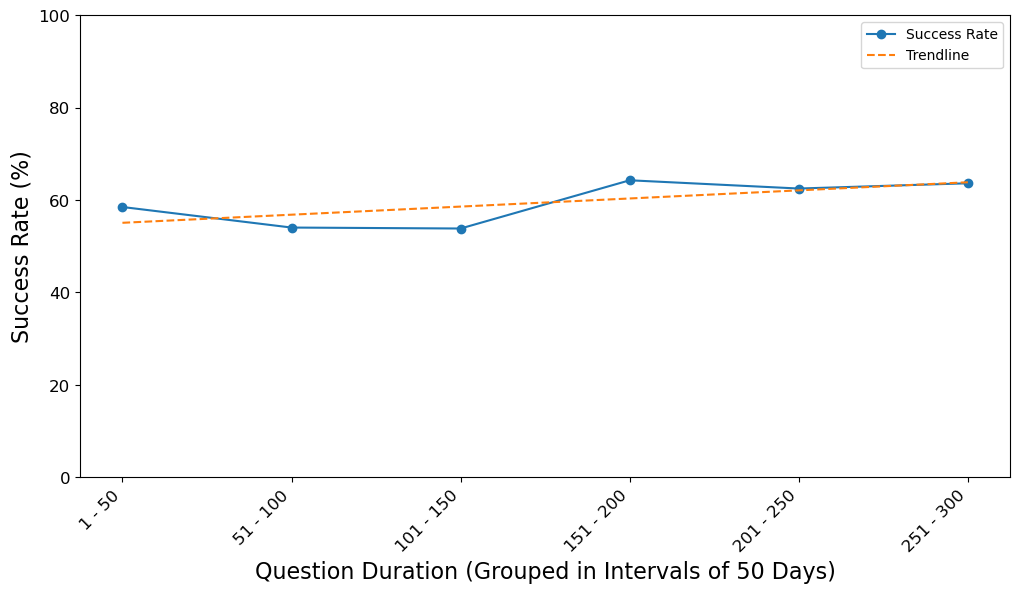}
        \caption{Input prompt: Q and B.}
        \label{fig:Alpaca-7B_qd_2}
    \end{subfigure}

    \vskip\baselineskip

    \begin{subfigure}[b]{0.23\textwidth}
        \centering
        \includegraphics[width=\textwidth]{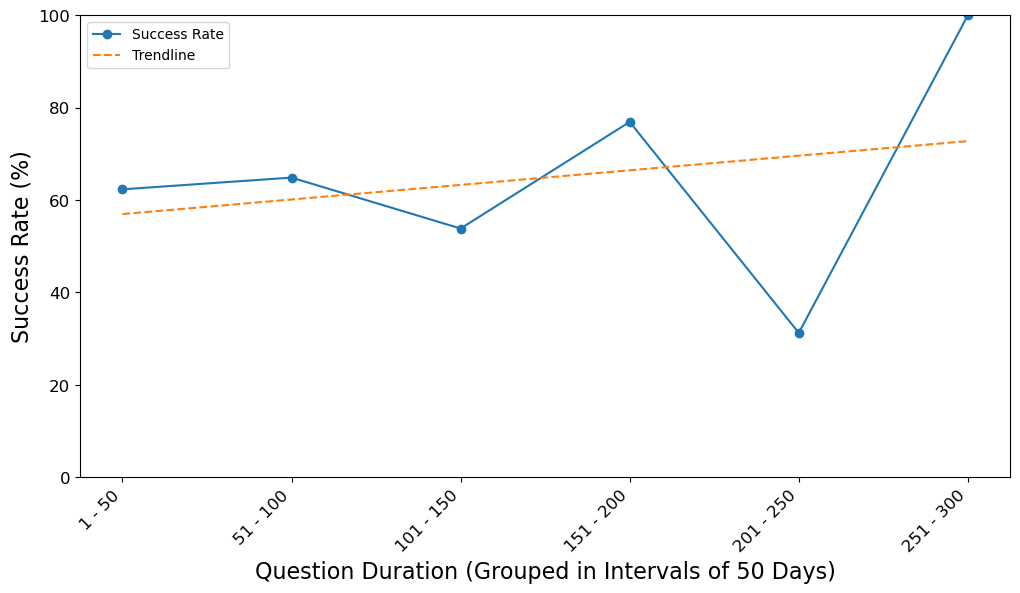}
        \caption{Input prompt: Q, B and NA.}
        \label{fig:Alpaca-7B_qd_3}
    \end{subfigure}
    \hfill
    \begin{subfigure}[b]{0.23\textwidth}
        \centering
        \includegraphics[width=\textwidth]{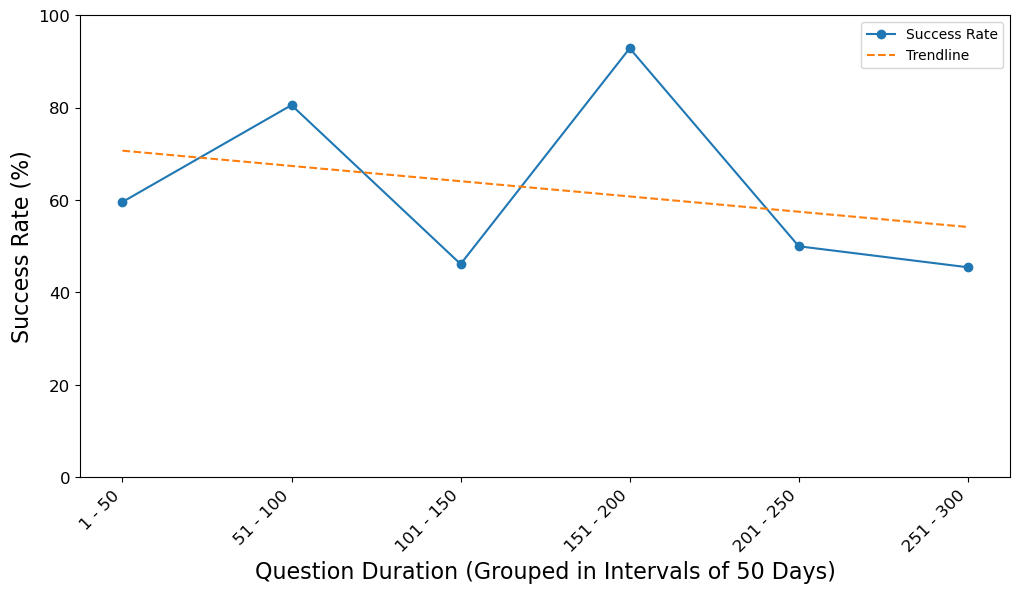}
        \caption{Input prompt: Q, B, NA and R.}
        \label{fig:Alpaca-7B_qd_4}
    \end{subfigure}

    \vskip\baselineskip

    \begin{subfigure}[b]{0.23\textwidth}
        \centering
        \includegraphics[width=\textwidth]{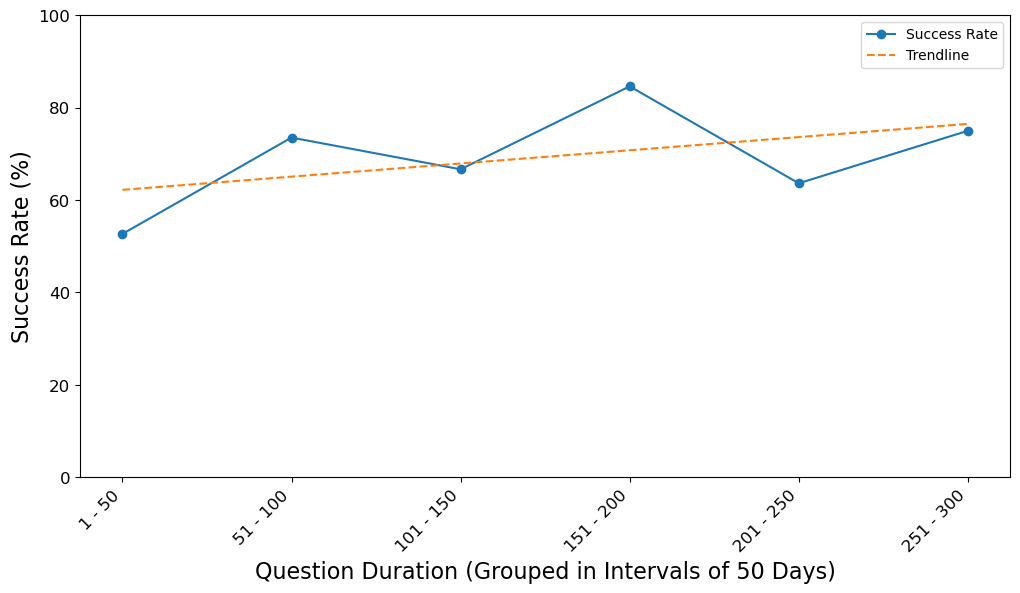}
        \caption{Input prompt: Q, B, NA, R and FS.}  
        \label{fig:Alpaca-7B_qd_5}
    \end{subfigure}
    
    \vskip\baselineskip
    \caption{Success rate across question duration, grouped in 50-day intervals (from 1 to 300 with a step of 50). Question forecasts were made by Alpaca-7B using various input prompts.}
    \label{fig:Alpaca-7B_question_duration_plots}
\end{figure}

\begin{figure}[H]
    \raggedright
    \includegraphics[width=0.35\textwidth]{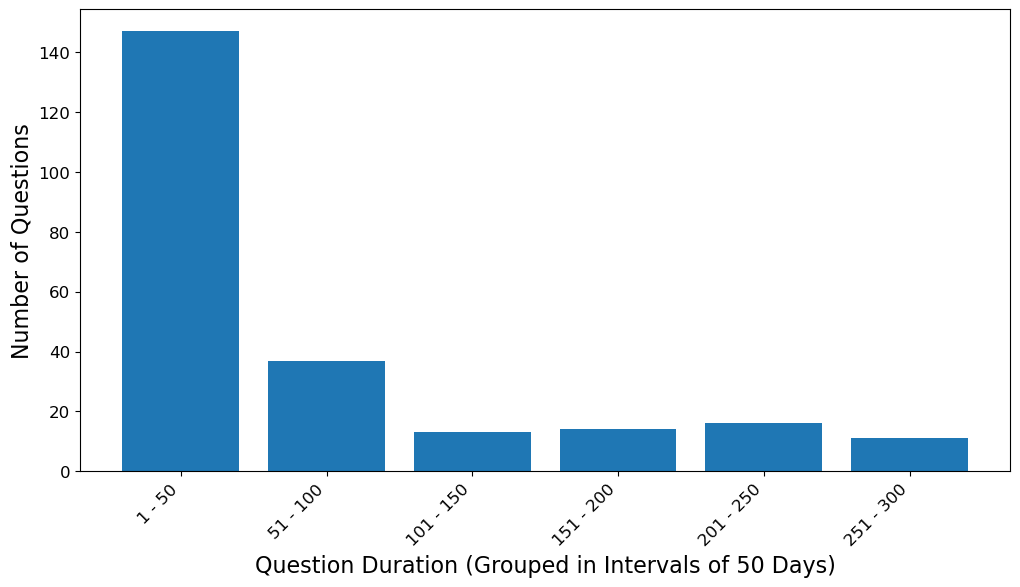}
    \caption{Question count across question duration using the original dataset (Alpaca-7B).}
    \label{fig:Alpaca-7B_qd_orig_numQ}
\end{figure}

\begin{figure}[H]
    \raggedright
    \includegraphics[width=0.42\textwidth]{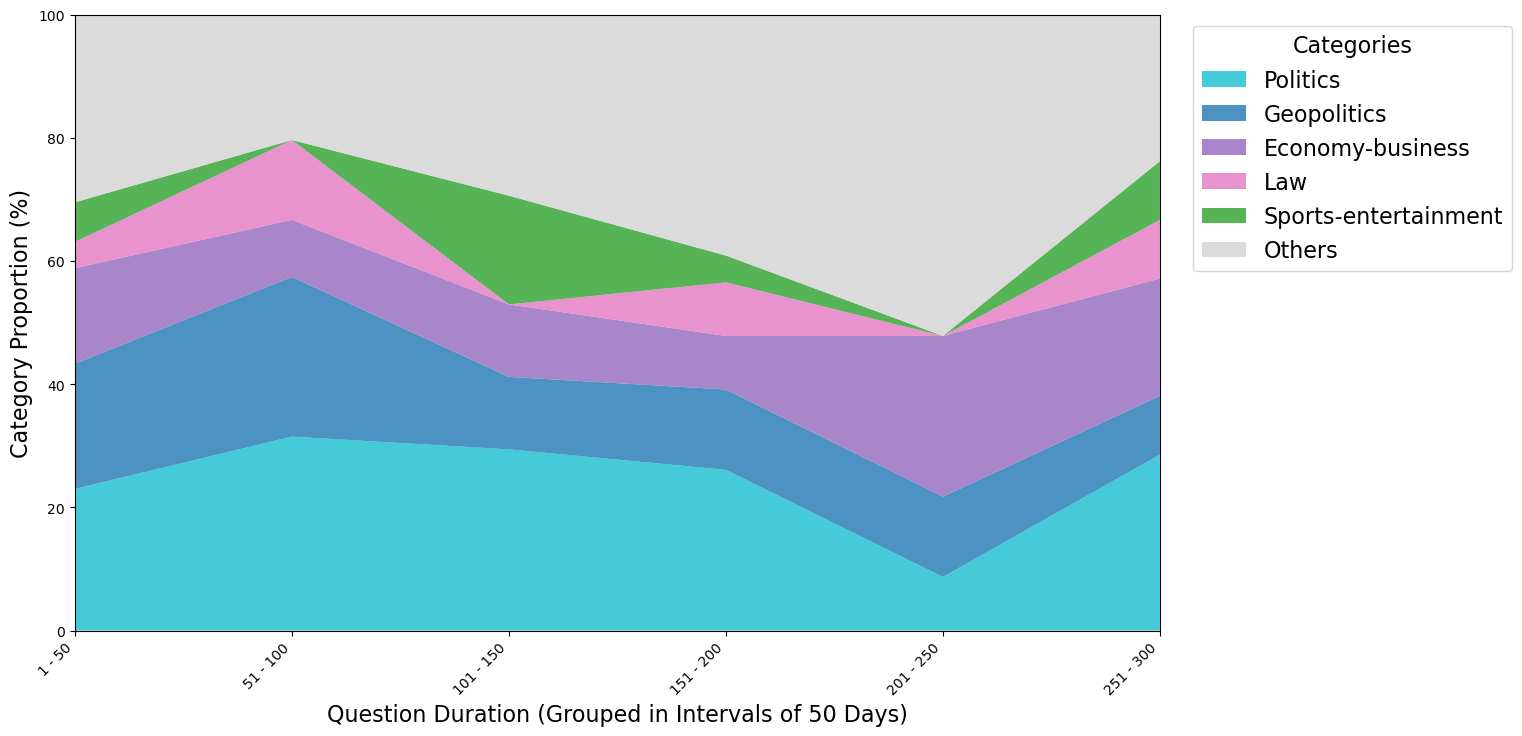}
    \caption{Ratio of the top five categories across question duration (Alpaca-7B).}
    \label{fig:Alpaca-7B_qd_orig_cat}
\end{figure}

\begin{figure}[H]
    \vskip\baselineskip
    \centering
    \begin{subfigure}[b]{0.23\textwidth}
        \centering
        \includegraphics[width=\textwidth]{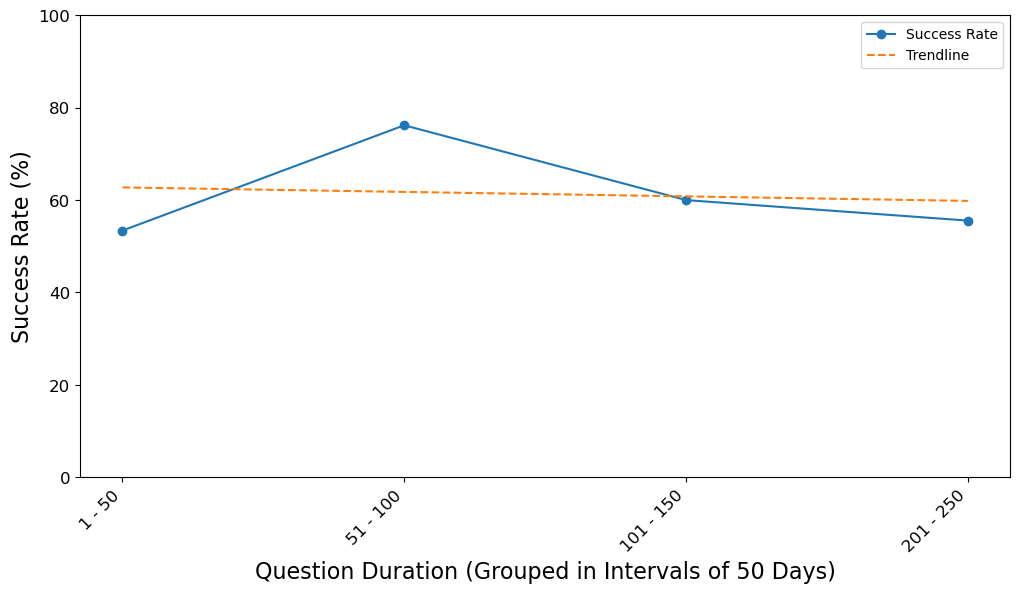}
        \caption{Input prompt: Q.}
        \label{fig:Llama2-13B_qd_1}
    \end{subfigure}
    \hfill
    \begin{subfigure}[b]{0.23\textwidth}
        \centering
        \includegraphics[width=\textwidth]{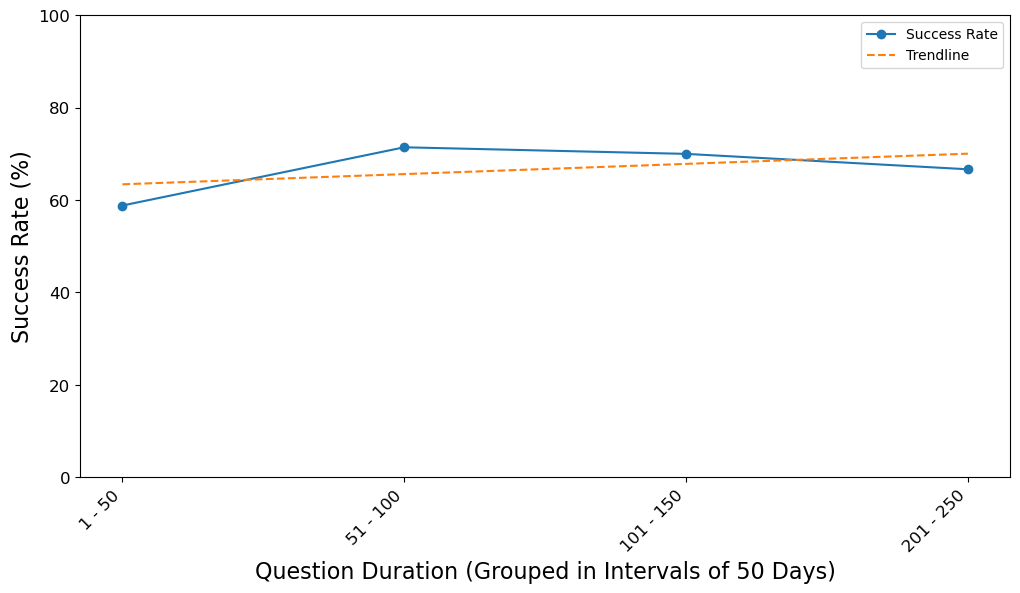}
        \caption{Input prompt: Q and B.}
        \label{fig:Llama2-13B_qd_2}
    \end{subfigure}
    \vskip\baselineskip

    \begin{subfigure}[b]{0.23\textwidth}
        \centering
        \includegraphics[width=\textwidth]{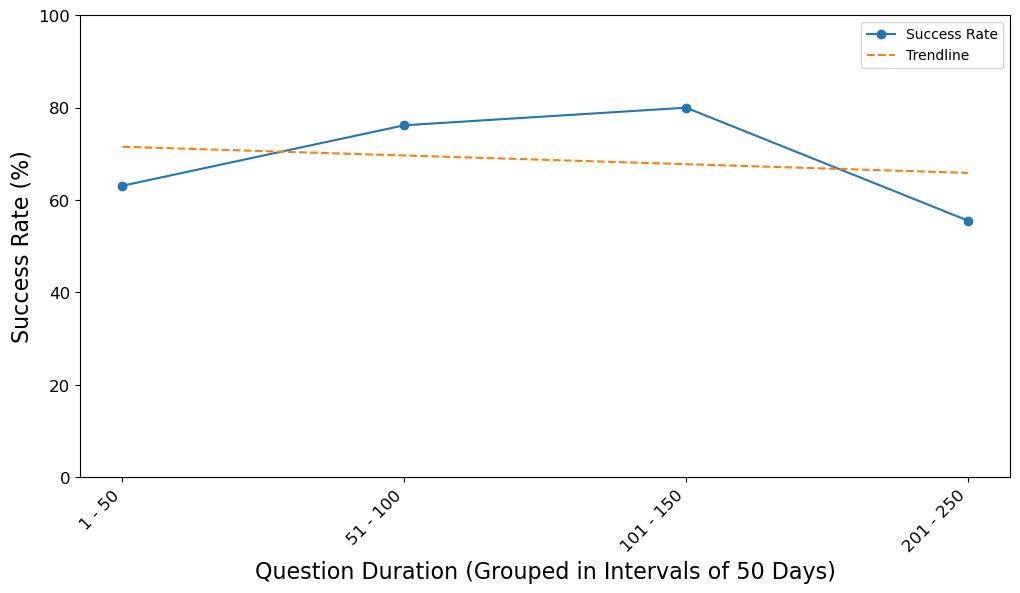}
        \caption{Input prompt: Q, B and NA.}
        \label{fig:Llama2-13B_qd_3}
    \end{subfigure}
    \hfill
    \begin{subfigure}[b]{0.23\textwidth}
        \centering
        \includegraphics[width=\textwidth]{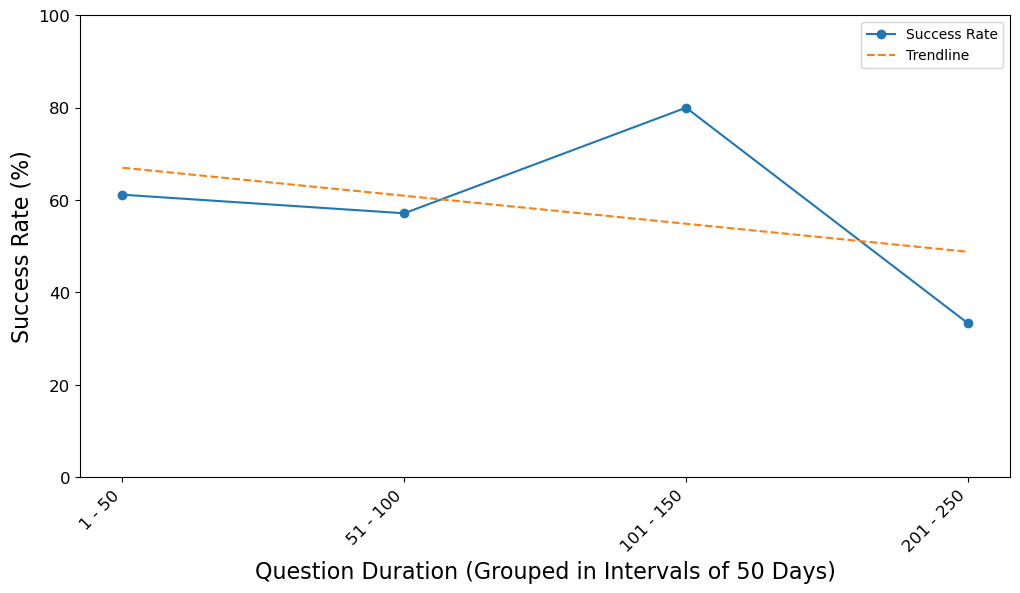}
        \caption{Input prompt: Q, B, NA and R.}
        \label{fig:Llama2-13B_qd_4}
    \end{subfigure}
    \vskip\baselineskip

    \begin{subfigure}[b]{0.23\textwidth}
        \centering
        \includegraphics[width=\textwidth]{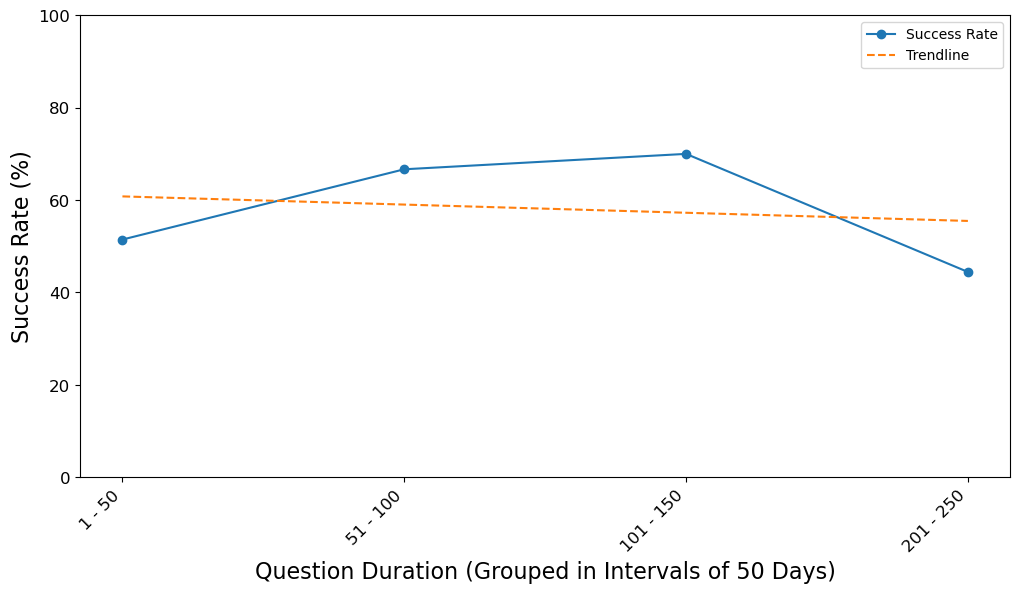}
        \caption{Input prompt: Q, B, NA, R and FS.}
        \label{fig:Llama2-13B_qd_5}
    \end{subfigure}
    
    \vskip\baselineskip
    \caption{Success rate across question duration, grouped in 50-day intervals (from 1 to 250 with a step of 50). Question forecasts were made by Llama2-13B-chat using various input prompts.}
    \label{fig:Llama2-13B_question_duration_plots}
\end{figure}

\begin{figure}[H]
    \raggedright
    \includegraphics[width=0.35\textwidth]{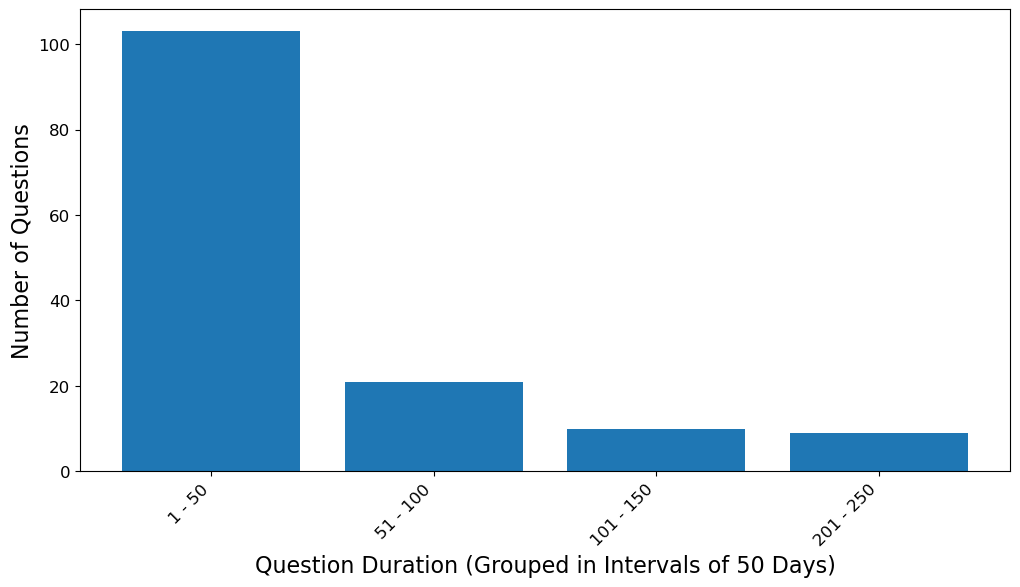}
    \caption{Question count across question duration (Llama2-13B-chat).}
    \label{fig:Llama2-13B_qd_orig_numQ}
\end{figure}

\begin{figure}[H]
    \raggedright
    \includegraphics[width=0.42\textwidth]{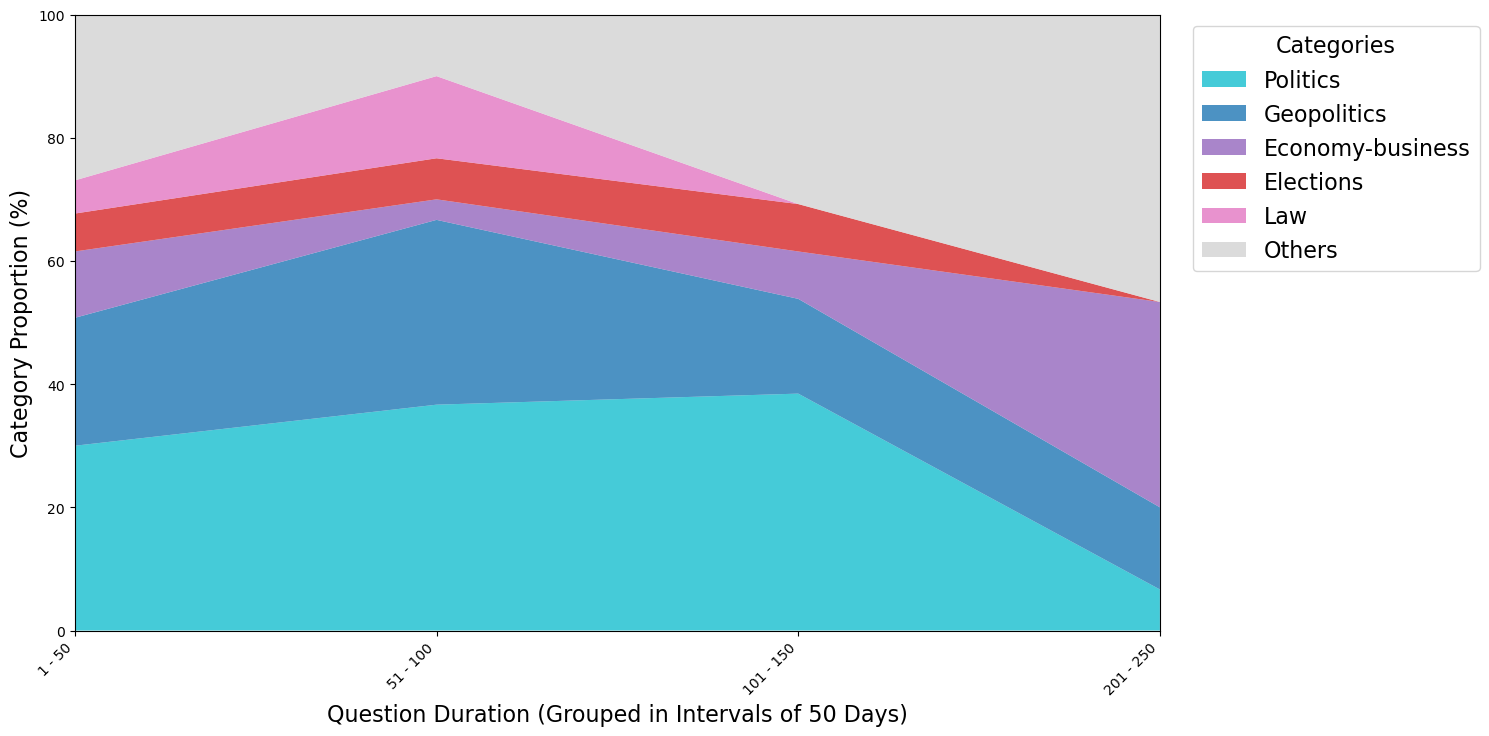}
    \caption{Ratio of the top five categories across question duration (Llama2-13B-chat).}
    \label{fig:Llama2-13B_qd_orig_cat}
\end{figure}

\end{document}